\documentclass[11pt,a4paper]{article}
\usepackage{authblk}
\usepackage[hyperref]{emnlp2020}
\usepackage{times}
\usepackage{latexsym}

\usepackage{microtype}
\usepackage[title]{appendix}

\aclfinalcopy 
\def\aclpaperid{2070}

\usepackage{graphicx}
\usepackage{subfigure}
\usepackage{booktabs}

\usepackage{hyperref}
\usepackage{amsmath}
\usepackage{amssymb}
\usepackage{mathtools}
\usepackage{amsthm}
\usepackage{algorithm}
\usepackage{algorithmic}
\usepackage[cal=zapfc, bb=ams, calscaled=1.2]{mathalfa}

\usepackage{lscape}
\usepackage{tikz}
\usepackage{pgfplots}

\usepackage[para]{threeparttable}
\usepackage{multirow}
\usepackage{array}
\usepackage{colortbl}

\usetikzlibrary{shapes}
\usetikzlibrary{calc}
\usetikzlibrary{positioning}
\usetikzlibrary{arrows}
\usetikzlibrary{bayesnet}
\usetikzlibrary{backgrounds}
\pgfplotsset{
	compat=1.3,
	box plot/.style={
		/pgfplots/.cd,
		black,
		only marks,
		mark=-,
		mark size=0.5em,
		/pgfplots/error bars/.cd,
		y dir=plus,
		y explicit,
	},
	box plot box/.style={
		/pgfplots/error bars/draw error bar/.code 2 args={%
			\draw  ##1 -- ++(0.5em,0pt) |- ##2 -- ++(-0.5em,0pt) |- ##1 -- cycle;   num
		},
		/pgfplots/table/.cd,
		y index=2,
		y error expr={\thisrowno{3}-\thisrowno{2}},
		/pgfplots/box plot
	},
	box plot top whisker/.style={
		/pgfplots/error bars/draw error bar/.code 2 args={%
			\pgfkeysgetvalue{/pgfplots/error bars/error mark}%
			{\pgfplotserrorbarsmark}%
			\pgfkeysgetvalue{/pgfplots/error bars/error mark options}%
			{\pgfplotserrorbarsmarkopts}%
			\path ##1 -- ##2;
		},
		/pgfplots/table/.cd,
		y index=4,
		y error expr={\thisrowno{2}-\thisrowno{4}},
		/pgfplots/box plot
	},
	box plot bottom whisker/.style={
		/pgfplots/error bars/draw error bar/.code 2 args={%
			\pgfkeysgetvalue{/pgfplots/error bars/error mark}%
			{\pgfplotserrorbarsmark}%
			\pgfkeysgetvalue{/pgfplots/error bars/error mark options}%
			{\pgfplotserrorbarsmarkopts}%
			\path ##1 -- ##2;
		},
		/pgfplots/table/.cd,
		y index=5,
		y error expr={\thisrowno{3}-\thisrowno{5}},
		/pgfplots/box plot
	},
	box plot median/.style={
		/pgfplots/box plot
	}
}

\renewcommand{\edge}[3][]{
    \foreach \x in {#2} {
        \foreach \y in {#3} {
            \message{#1};
            \path (\x) edge [->, >={triangle 45}, #1] (\y);
        };
    };
}

\renewcommand{\plate}[4][]{ 
    \node[wrap=#3] (#2-wrap) {};
    \node[plate caption=#2-wrap] (#2-caption) {#4};
    \node[plate=(#2-wrap)(#2-caption), #1] (#2) {};
}

\DeclarePairedDelimiterX{\curlyx}[1]{\{}{\}}{#1}
\DeclarePairedDelimiterX{\roundx}[1]{(}{)}{#1}
\DeclarePairedDelimiterX{\squarex}[1]{[}{]}{#1}
\DeclarePairedDelimiterX{\gaussianx}[2]{(}{)}{#1,#2}
\DeclarePairedDelimiterX{\roundcondx}[2]{(}{)}{#1\delimsize\vert#2}
\DeclarePairedDelimiterX{\kldx}[2]{(}{)}{#1\delimsize\|#2}

\DeclareMathOperator*{\decoder}{\mathcal{D}}
\DeclareMathOperator*{\argmax}{arg\,max}

\DeclareMathOperator*{\softmax}{softmax}
\DeclareMathOperator*{\softplus}{softplus}
\DeclareMathOperator*{\lstm}{LSTM}

\DeclareMathOperator*{\seqencoder}{\mathcal{E}}
\DeclareMathOperator*{\ctxencoder}{\mathcal{C}}
\DeclareRobustCommand\sampleline[1]{
	\tikz\draw[#1] (0,0) (0,\the\dimexpr\fontdimen22\textfont2\relax)
	-- (1em,\the\dimexpr\fontdimen22\textfont2\relax);
}
\newcommand{\boldx}{\mathbf{x}}

\newcommand{\boldz}{\mathbf{z}}
\newcommand{\boldc}{\mathbf{c}}
\newcommand{\boldu}{\mathbf{u}}
\newcommand{\boldr}{\mathbf{r}}
\newcommand{\boldg}{\mathbf{g}}
\newcommand{\bolds}{\mathbf{s}}
\newcommand{\boldv}{\mathbf{v}}
\newcommand{\boldh}{\mathbf{h}}
\newcommand{\boldmu}{\mathbf{\mu}}
\newcommand{\boldsigma}{\mathbf{\sigma}}

\newcommand{\zspeaker}{\boldz^{\roundx{r}}}
\newcommand{\zgoal}{\boldz^{\roundx{g}}}
\newcommand{\zstate}{\boldz^{\roundx{s}}}
\newcommand{\zutt}{\boldz^{\roundx{u}}}
\newcommand{\zconv}{\boldz^{\roundx{c}}}
\newcommand{\hspeaker}{\boldh^{\roundx{r}}}
\newcommand{\hgoal}{\boldh^{\roundx{g}}}
\newcommand{\hstate}{\boldh^{\roundx{s}}}
\newcommand{\hutt}{\boldh^{\roundx{u}}}
\newcommand{\muspeaker}{\boldmu^{\roundx{r}}}
\newcommand{\mugoal}{\boldmu^{\roundx{g}}}
\newcommand{\mustate}{\boldmu^{\roundx{s}}}
\newcommand{\muutt}{\boldmu^{\roundx{u}}}
\newcommand{\sigmaspeaker}{\boldsigma^{\roundx{r}}}
\newcommand{\sigmagoal}{\boldsigma^{\roundx{g}}}
\newcommand{\sigmastate}{\boldsigma^{\roundx{s}}}
\newcommand{\sigmautt}{\boldsigma^{\roundx{u}}}

\newcommand{\identity}{\mathbf{I}}
\newcommand{\posteriorapp}{q_\phi\roundcondx}
\newcommand{\posteriormodel}{p_\theta\roundcondx}
\newcommand{\gaussian}{\mathcal{N}\gaussianx}
\newcommand{\kld}{D_{\text{KL}}\kldx}
\newcommand{\super}{\textsuperscript}
\newcommand{\weakline}{
	\noalign{\global\arrayrulewidth=0.25\arrayrulewidth}
    \arrayrulecolor{lightgray}
    \\ [-1.8ex] \hline \\ [-1.8ex]
	\noalign{\global\arrayrulewidth=4.0\arrayrulewidth}
	\arrayrulecolor{black}
}

\title{Variational Hierarchical Dialog Autoencoder for \\
       Dialog State Tracking Data Augmentation}

\author[1]{\textbf{Kang Min Yoo}}
\author[1]{\textbf{Hanbit Lee}}
\author[2]{\textbf{Franck Dernoncourt}}
\author[2]{\textbf{Trung Bui}}
\author[2]{\authorcr\textbf{Walter Chang}}
\author[1]{\textbf{Sang-goo Lee}}
\affil[1]{Seoul National University, Seoul, Korea}
\affil[2]{Adobe Research, San Jose, CA, USA}
\affil[ ]{\texttt{\{kangminyoo,skcheon,sglee\}@europa.snu.ac.kr}}
\affil[ ]{\texttt{\{dernonco,bui,wachang\}@adobe.com}}

\date{}

\begin{document}
\maketitle
\begin{abstract}
	Recent works have shown that generative data augmentation, where synthetic 
	    samples generated from deep generative models complement 
	    the training dataset, benefit NLP tasks.
	In this work, we extend this approach to the task of dialog state tracking 
		for goal-oriented dialogs.
    Due to the inherent hierarchical structure of goal-oriented dialogs over 
        utterances and related annotations, the deep generative model must be 
        capable of capturing the coherence among different hierarchies and types 
        of dialog features.
	We propose the Variational Hierarchical Dialog Autoencoder (VHDA) 
		for modeling the complete aspects of goal-oriented dialogs, including 
		linguistic features and underlying structured annotations, namely
		speaker information, dialog acts, and goals. 
	The proposed architecture is designed to model each aspect of goal-oriented 
		dialogs using inter-connected latent variables and learns to generate coherent goal-oriented dialogs from the latent spaces.
	To overcome training issues that arise from training complex variational 
		models, we propose appropriate training strategies.
    Experiments on various dialog datasets show that our model improves the 
        downstream dialog trackers' robustness via generative data augmentation.
	We also discover additional benefits of our unified approach to modeling 
		goal-oriented dialogs -- dialog response generation and user simulation, 
		where our model outperforms previous strong baselines.
\end{abstract}

\section{Introduction}

Data augmentation, a technique that augments the training set with 
	label-preserving synthetic samples, is commonly employed in modern machine
	learning approaches.
It has been used extensively in visual learning pipelines
	\cite{shorten2019survey} but less frequently for NLP 
	tasks due to the lack of well-established techniques in the area.
While some notable work exists in 
	text classification \cite{zhang2015character}, spoken language 
	understanding \cite{yoo2019data}, and machine translation
	\cite{fadaee2017data}, we still lack the full understanding of 
	utilizing generative models for text augmentation.

Ideally, a data augmentation technique for supervised tasks must
	synthesize \textit{distribution-preserving} and 
	\textit{sufficiently realistic} samples.
Current approaches for data augmentation in NLP tasks mostly revolve around
	thesaurus data augmentation \cite{zhang2015character}, in
	which words that belong to the same semantic role are substituted with one
	another using a preconstructed lexicon, and noisy data augmentation
	\cite{wei2019eda} where random editing operations create perturbations in
	the language space.
Thesaurus data augmentation requires a set of handcrafted semantic dictionaries, 
	which are costly
	to build and maintain, whereas noisy data augmentation does not synthesize
	sufficiently realistic samples.
The recent trend \cite{hu2017toward,yoo2019data,shin2019utterance} gravitates 
	towards \textit{generative data augmentation}
	(GDA), a class of techniques that leverage deep generative models 
	such as VAEs to delegate the automatic discovery of novel class-preserving samples to machine learning.
In this work, we explore GDA in the context of dialog modeling and contextual
	understanding.

Goal-oriented dialogs occur between a user and a system that communicates
	verbally to accomplish the user's goals 
	(\autoref{tab:vhda-interpolation-2}).
However, because the user's goals and the system's possible actions are not 
	transparent to each other, both parties must rely on verbal communications 
	to infer and take appropriate actions to resolve the goals.
Dialog state tracker is a core component of such systems, 
	enabling it to track the dialog's latest status \cite{henderson2014second}.
A dialog state typically consists of \texttt{inform} and \texttt{request}
	types of slot values.
For example, a user may verbally refer to a previously mentioned food type as 
    the preferred one - e.g., Asian (\verb|inform(food=asian)|).
Given the user utterance and historical turns, the state tracker must infer
	the user's current goals.
As such, we can view dialog state tracking as a sparse sequential 
	multi-class classification problem.
Modeling goal-oriented dialogs for GDA requires a novel approach 
	that simultaneously solves state tracking, user simulation 
	\cite{schatzmann2007agenda}, and utterance generation.

Various deep models exist for modeling dialogs.
The Markov approach \cite{serban2017hierarchical} employs
	a sequence-to-sequence variational autoencoder (VAE) 
	\cite{kingma2013auto} structure to predict the next utterance given
	a deterministic context representation, while the holistic approach
	\cite{park2018hierarchical} utilizes a set of global
	latent variables to encode the entire dialog,
	improving the awareness in general dialog structures.
However, current approaches are limited to linguistic features.
Recently, \citet{bak2019variational} proposed a hierarchical VAE structure
	that incorporates the speaker's information, but we have yet
	to explore a universal approach for encompassing fundamental aspects of 
	goal-oriented dialogs.
Such a unified model capable of disentangling latents into specific dialog 
    aspects can increase the modeling efficiency and enable interesting 
    extensions based on the fine-grained controllability.

This paper proposes a 
	novel multi-level hierarchical and recurrent VAE structure
	called Variational Hierarchical Dialog Autoencoder (VHDA).
Our model enables modeling all aspects (speaker information, goals, dialog 
	acts, utterances, and general dialog flow) of goal-oriented dialogs in a disentangled manner by assigning latents to each aspect.
However, complex and autoregressive VAEs are known to suffer from the risk of
	\textit{inference collapse} \cite{cremer2018inference}, in which 
	the model converges to a local optimum where the generator network
	neglects the latents, reducing the generation controllability.
To mitigate the issue, we devise two simple but effective training strategies.

Our contributions are summarized as follows. 
\begin{enumerate}
    \item We propose a novel deep latent model for modeling dialog utterances and 
	their relationships with the goal-oriented annotations.
	We show that the strong level of coherence and accuracy displayed by 
	the model allows it to be used for augmenting dialog state tracking
	datasets.
	\item Leveraging the model's generation capabilities, we show that generative 
    data augmentation is attainable even for the complex dialog-related 
    tasks that pertain to both hierarchical and sequential annotations.
    \item We propose simple but effective training policies for our VAE-based 
    model, which have applications in other similar VAE structures.
\end{enumerate}

The code for reproducing this paper is available at github
    \footnote{\url{https://github.com/kaniblu/vhda}}.
    
\section{Background and Related Work}
\label{sec:vhda-background}

\textbf{Dialog State Tracking. }
Dialog state tracking (DST) predicts the user’s current goals and dialog 
    acts, given the dialog context.
Historically, DST models have gradually evolved from hand-crafted finite-state 
	automata and multi-stage models 
	\cite{dybkjaer2008recent, thomson2010bayesian, wang2013simple} to 
    end-to-end models that
    directly predict dialog states from dialog features
    \cite{zilka2015incremental, mrkvsic2017neural, zhong2018global,
		  nouri2018toward}.

Among the proposed models, Neural Belief Tracker (NBT) \cite{mrkvsic2017neural} 
    decreases reliance on handcrafted semantic dictionaries by 
	reformulating the classification problem.
Global-locally Self-attentive Dialog tracker (GLAD) \cite{zhong2018global} 
	introduces global modules for sharing parameters across slots and local 
	modules, allowing the learning of slot-specific feature representations.
Globally-Conditioned Encoder (GCE) \cite{nouri2018toward} improves further 
	by forgoing the separation of global and local modules, allowing the unified 
	module to take slot embeddings for distinction. 
Recently, dialog state trackers based on pre-trained language models
	have demonstrated their strong performance in many DST tasks
	\cite{wu2019transferable, kim2019efficient, hosseini2020simple}.
While the utilization of large-scale pre-trained language models is not within 
    our scope, we wish to explore further concerning the recent advances in the area.

\noindent\textbf{Conversation Modeling. }
While the previous approaches for hierarchical dialog modeling relate to 
    the Markov assumption \cite{serban2017hierarchical}, 
	recent approaches have geared towards utilizing global latent variables 
	for representing the holistic dialog structure
	\cite{park2018hierarchical,gu2018dialogwae,bak2019variational}, 
	which helps in preserving long-term dependencies and total semantics.
In this work, we employ global latent variables to maximize the effectiveness
	in preserving dialog semantics for data augmentation.

\noindent\textbf{Data Augmentation.}
Transformation-based data augmentation is popular in vision learning
    \cite{shorten2019survey} and speech signal processing \cite{ko2015audio},
    while thesaurus and noisy data augmentation techniques are more common for text.
    \cite{zhang2015character, wei2019eda}.
Recently, generative data augmentation (GDA), augmenting data gather from samples
    generated from fine-tuned deep generative models, 
    have gained traction in several NLP tasks
    \cite{hu2017toward, hou2018sequence, yoo2019data, shin2019utterance}.
GDA can be seen as a form of unsupervised data augmentation, delegating
    the automatic discovery of novel data to machine learning without 
    injecting external knowledge or data sources.
While most works utilize VAE for the generative model,
    some works achieved a similar effect without employing 
    variational inference \cite{kurata2016leveraging, hou2018sequence}.
In contrast to unsupervised data augmentation, 
    another line of work has explored self-supervision mechanisms to 
    fine-tune the generators for specific tasks 
    \cite{tran2017bayesian, antoniou2017data, cubuk2018autoaugment}.
Recent work proposed a reinforcement learning-based noisy data augmentation 
    framework for state tracking \cite{yin2019dialog}.
Our work belongs to the family of unsupervised GDA, which can incorporate 
    self-supervision mechanisms.
We wish to explore further in this regard.

\section{Proposed Model}
\label{sec:vhda-model}

\begin{figure}
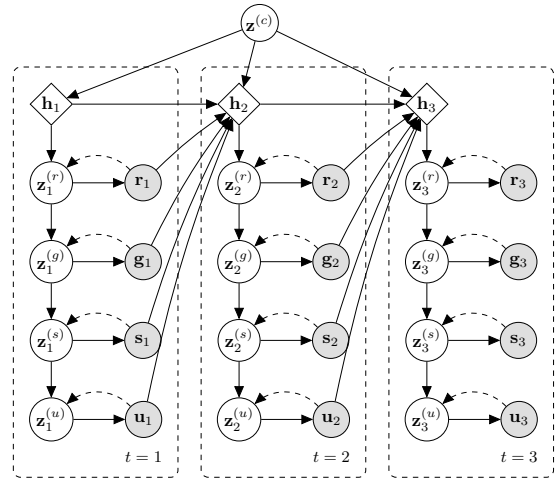

	\centering
	\scalebox{0.68}{\tikz{
		\node [latent, diamond] (context1) {$\boldh_1$};
		\node [latent, below=0.7cm of context1] (z_speaker1) {$\zspeaker_1$};
		\node [latent, below=0.7cm of z_speaker1] (z_goal1) {$\zgoal_1$};
		\node [latent, below=0.7cm of z_goal1] (z_state1) {$\zstate_1$};
		\node [latent, below=0.7cm of z_state1] (z_utt1) {$\zutt_1$};
		\node [obs, right=of z_speaker1] (obs_speaker1) {$\boldr_1$};
		\node [obs, right=of z_goal1] (obs_goal1) {$\boldg_1$};
		\node [obs, right=of z_state1] (obs_state1) {$\bolds_1$};
		\node [obs, right=of z_utt1] (obs_utt1) {$\boldu_1$};
		
		\node [latent, diamond, right=of context1, xshift=1.8cm] 
			(context2) {$\boldh_2$};
		\node [latent, below=0.7cm of context2] (z_speaker2) {$\zspeaker_2$};
		\node [latent, below=0.7cm of z_speaker2] (z_goal2) {$\zgoal_2$};
		\node [latent, below=0.7cm of z_goal2] (z_state2) {$\zstate_2$};
		\node [latent, below=0.7cm of z_state2] (z_utt2) {$\zutt_2$};
		\node [obs, right=of z_speaker2] (obs_speaker2) {$\boldr_2$};
		\node [obs, right=of z_goal2] (obs_goal2) {$\boldg_2$};
		\node [obs, right=of z_state2] (obs_state2) {$\bolds_2$};
		\node [obs, right=of z_utt2] (obs_utt2) {$\boldu_2$};
		
		\node [latent, diamond, right=of context2, xshift=1.8cm] 
			(context3) {$\boldh_3$};
		\node [latent, below=0.7cm of context3] (z_speaker3) {$\zspeaker_3$};
		\node [latent, below=0.7cm of z_speaker3] (z_goal3) {$\zgoal_3$};
		\node [latent, below=0.7cm of z_goal3] (z_state3) {$\zstate_3$};
		\node [latent, below=0.7cm of z_state3] (z_utt3) {$\zutt_3$};
		\node [obs, right=of z_speaker3] (obs_speaker3) {$\boldr_3$};
		\node [obs, right=of z_goal3] (obs_goal3) {$\boldg_3$};
		\node [obs, right=of z_state3] (obs_state3) {$\bolds_3$};
		\node [obs, right=of z_utt3] (obs_utt3) {$\boldu_3$};
		
		\node [latent, above=0.8cm of context2, xshift=0.4cm] 
			(z_conv) {$\zconv$};
		
		\plate [dashed, inner sep=0.3cm] {group1} {
			(context1)
			(z_goal1)(z_state1)(z_speaker1)(z_utt1)
			(obs_goal1)(obs_state1)(obs_speaker1)(obs_utt1)
		} {$t = 1$};
		\plate [dashed, inner sep=0.3cm] {group2} {
			(context2)
			(z_goal2)(z_state2)(z_speaker2)(z_utt2)
			(obs_goal2)(obs_state2)(obs_speaker2)(obs_utt2)
		} {$t = 2$};
		\plate [dashed, inner sep=0.3cm] {group3} {
			(context3)
			(z_goal3)(z_state3)(z_speaker3)(z_utt3)
			(obs_goal3)(obs_state3)(obs_speaker3)(obs_utt3)
		} {$t = 3$};

		\edge {context1} {z_speaker1, context2};
		\edge {z_speaker1} {z_goal1, obs_speaker1};
		\edge {z_goal1} {z_state1, obs_goal1};
		\edge {z_state1} {z_utt1, obs_state1};
		\edge {z_utt1} {obs_utt1};
		\edge [bend left=5] 
			{obs_speaker1, obs_goal1, obs_state1, obs_utt1} {context2};
		
		\edge [dashed, bend right=45] {obs_speaker1} {z_speaker1};
		\edge [dashed, bend right=45] {obs_goal1} {z_goal1};
		\edge [dashed, bend right=45] {obs_state1} {z_state1};
		\edge [dashed, bend right=45] {obs_utt1} {z_utt1};

		\edge {context2} 
			{z_speaker2, context3};
		\edge {z_speaker2} {z_goal2, obs_speaker2};
		\edge {z_goal2} {z_state2, obs_goal2};
		\edge {z_state2} {z_utt2, obs_state2};
		\edge {z_utt2} {obs_utt2};
		\edge [bend left=5] 
			{obs_speaker2, obs_goal2, obs_state2, obs_utt2} {context3};

		\edge [dashed, bend right=45] {obs_speaker2} {z_speaker2};
		\edge [dashed, bend right=45] {obs_goal2} {z_goal2};
		\edge [dashed, bend right=45] {obs_state2} {z_state2};
		\edge [dashed, bend right=45] {obs_utt2} {z_utt2};

		\edge {context3} {z_speaker3};
		\edge {z_speaker3} {z_goal3, obs_speaker3};
		\edge {z_goal3} {z_state3, obs_goal3};
		\edge {z_state3} {z_utt3, obs_state3};
		\edge {z_utt3} {obs_utt3};
		
		\edge {z_conv} {context1, context2, context3};

		\edge [dashed, bend right=45] {obs_speaker3} {z_speaker3};
		\edge [dashed, bend right=45] {obs_goal3} {z_goal3};
		\edge [dashed, bend right=45] {obs_state3} {z_state3};
		\edge [dashed, bend right=45] {obs_utt3} {z_utt3};
	}}
	\caption{Graphical representation of VHDA. 
			 Solid and dashed arrows represent generation and recognition 
			 respectively.}
	\label{fig:vhda}
\end{figure}

This section describes VHDA, our latent variable model for generating 
	goal-oriented dialog datasets.
We first introduce a set of notations for describing core concepts.

\subsection{Notations}

A dialog dataset $\mathbb{D}$ is a set of $N$ i.i.d 
	samples $\curlyx*{\boldc_1, \ldots, \boldc_N}$, where
	each $\boldc$ is a sequence of turns 
	$\roundx{\boldv_1, \ldots, \boldv_T}$.
Each goal-oriented dialog turn $\boldv$ is a tuple of speaker information 
	$\boldr$, the speaker's goals $\boldg$, dialog state $\bolds$, and the 
	speaker's utterance $\boldu$: 
	$\boldv = \roundx{\boldr, \boldg, \bolds, \boldu}$.
Each utterance $\boldu$ is a sequence of words 
	$\roundx{w_1, \ldots, w_{|\boldu|}}$.
Goals $\boldg$ or a dialog state $\bolds$ is defined 
	as a set of the smallest unit of dialog act specification $a$
	\cite{henderson2014second}, which is a tuple of dialog act, slot and value
	defined over the space of $\mathcal{T}$, $\mathcal{S}$, and $\mathcal{V}$: 
	$\boldg = \curlyx*{a_1, \ldots, a_{|\boldg|}}$,
	$\bolds = \curlyx*{a_1, \ldots, a_{|\bolds|}}$, 
	where $a_i \in \mathcal{A} = 
		   \roundx{\mathcal{T}, \mathcal{S}, \mathcal{V}}$.
A dialog act specification is represented as 
	\verb|<act>|(\verb|<slot>|=\verb|<value>|).

\subsection{VHCR}

Given a conversation $\boldc$, Variational Hierarchical Conversational RNN 
	(VHCR) \cite{park2018hierarchical} models
	the holistic features of the conversation and
	the individual utterances $\boldu$ using a hierarchical and 
	recurrent VAE model.
The model introduces global-level latent variables $\zconv$ for encoding 
	the high-level dialog structure and, at each turn $t$, local-level latent 
	variables $\zutt_t$ responsible for encoding and generating the utterance at 
	turn $t$.
The local latent variables $\zutt$ conditionally depends on
	$\zconv$ and previous observations, forming a hierarchical structure with
	the global latents.
Furthermore, hidden variables $\boldh_t$, which are conditionally
	dependent on the global information and the hidden variables from the 
	previous step $\boldh_{t-1}$, facilitate the latent inference.

\subsection{VHDA}

We propose Variational Hierarchical Dialog Autoencoder 
	(VHDA) to generate dialogs and their underlying dialog annotations 
	simultaneously (\autoref{fig:vhda}).
Like VHCR, we employ a hierarchical VAE structure to capture
	holistic dialog semantics using the conversation latent variables $\zconv$.
Our model incorporates full dialog features using turn-level latents
	$\zspeaker$ (speaker), $\zgoal$ (goal), $\zstate$ (dialog state), and 
	$\zutt$ (utterance), motivated by speech act theory 
	\cite{searle1980speech}.
Specifically, at a given dialog turn, 
	the information about the speaker, 
	the speaker's goals, the speaker's turn-level dialog acts, and the
	utterance all cumulatively determine one after the other in that order.

VHDA consists of multiple encoder and decoder modules, each 
	responsible for encoding or generating a particular dialog 
	feature.
The encoders share the identical sequence-encoding architecture 
	described as follows.

\noindent\textbf{Sequence Encoder Architecture}. 
Given a sequence of variable number of elements 
	$\mathbf{X} = \squarex{\boldx_1; \ldots; \boldx_n}^\intercal
				  \in \mathbb{R}^{n \times d}$, where $n$ is the number of 
	elements, the goal of a sequence 
	encoder is to extract a fixed-size representation $\boldh \in \mathbb{R}^d$, 
	where $d$ is the dimensionality of the hidden representation.
For our implementation, we employ the self-attention mechanism over 
	hidden outputs of bidirectional LSTM \cite{hochreiter1997long} cells 
	produced from the input sequence.
We also allow the attention mechanism to be optionally queried by $\mathbf{Q}$, 
	enabling the sequence to depend on external 
	conditions, such as using the dialog context to attend over an utterance:
\begin{align}
	\mathbf{H} & = \squarex{\overrightarrow{\lstm}\roundx{\mathbf{X}}; 
							 \overleftarrow{\lstm}\roundx{\mathbf{X}}}
				   \in \mathbb{R}^{n\times d} \nonumber \\
	\mathbf{a} & = \softmax\roundx{
		\squarex{\mathbf{H}; \mathbf{Q}} \mathbf{w} + b} 
		\in \mathbb{R}^n \nonumber \\
	\mathbf{h} & = \mathbf{H}^\intercal \mathbf{a} \in \mathbb{R}^{d}. \nonumber
\end{align}
Here, $\mathbf{Q} \in \mathbb{R}^{n \times d_q}$ is a collection of query 
	vectors of size $d_q$ where each vector corresponds to one element in the 
	sequence; $\mathbf{w} \in \mathbb{R}^{d + d_q}$ and 
	$b \in \mathbb{R}$ are learnable parameters.
We encapsulate the above operations with the following notation:
\begin{align}
	\seqencoder: \mathbb{R}^{n \times d} ( \times \mathbb{R}^{n \times d_q} )
		\rightarrow \mathbb{R}^{d}. \nonumber
\end{align}
Our model utilizes the $\seqencoder$ structure for encoding dialog 
	features of variable lengths.

\noindent\textbf{Encoder Networks}. 
Based on the $\seqencoder$ architecture, feature encoders are 
	responsible for encoding dialog features from their respective raw feature 
	spaces to hidden representations.
For goals and turn states, the encoding consists of two steps.
Initially, the multi-purpose dialog act encoder $\seqencoder^{(a)}$ 
	processes each dialog act triple of the goals $a^{(g)} \in \boldg$ and 
	turn states $a^{(s)} \in \bolds$ into a fixed-size 
	representation $\boldh^{(a)} \in \mathbb{R}^{d^{(a)}}$.
The encoder treats the dialog act triples as sequences of tokens.
Subsequently, the goal encoder and the turn state encoder process those 
    dialog act representations to produce goal representations and turn state
    representations, respectively:
\begin{align}
	\hgoal & = \textstyle \seqencoder^{(g)} \roundx{\squarex{
		\seqencoder^{(a)}\roundx{a^{(g)}_1}; \ldots; 
		\seqencoder^{(a)}\roundx{a^{(g)}_{|\boldg|}}
	}} \nonumber \\
	\hstate & = \textstyle \seqencoder^{(s)} \roundx{\squarex{
		\seqencoder^{(a)}\roundx{a^{(s)}_1}; \ldots; 
		\seqencoder^{(a)}\roundx{a^{(s)}_{|\bolds|}}
	}}. \nonumber
\end{align}
Note that, as the model is sensitive to the order of the dialog acts, 
    we randomize the order during training to prevent overfitting.
The utterances are encoded using the utterance encoder from the word 
	embeddings space:
	$\hutt = \textstyle \seqencoder^{(u)} \roundx{\squarex{
		\mathbf{w}_1; \ldots; \mathbf{w}_{|u|}
	}}$, while the entire conversation is encoded by the conversation encoder
	from the encoded utterance vectors:
	$\mathbf{h}^{(c)} = \textstyle \seqencoder^{(c)} \roundx{\squarex{
		\hutt_1; \ldots; \hutt_T
	}}$.
All sequence encoders mentioned above depend on the global latent variables 
	$\zconv$ via the query vector.
For the speaker information, we use the speaker embedding matrix
    $\mathbf{W}^{\roundx{r}} \in \mathbb{R}^{n^{\roundx{r}} 
    \times d^{\roundx{r}}}$ to encode the speaker vectors $\hspeaker$, where 
	$n^{\roundx{r}}$ is the number of participants and $d^{\roundx{r}}$ is 
	the embedding size.

\noindent\textbf{Main Architecture}.
At the top level, our architecture consists of five $\seqencoder$ encoders, 
    a context encoder $\ctxencoder$, and four types of decoder $\decoder$.
The context encoder $\ctxencoder$ is different from the other encoders,
	as it does \textit{not} utilize the 
	bidirectional $\seqencoder$ architecture but a uni-directional LSTM cell.
The four decoders $\decoder^{\roundx{r}}$, $\decoder^{\roundx{g}}$, 
	$\decoder^{\roundx{s}}$, and $\decoder^{\roundx{u}}$ generate respective
	dialog features.

$\ctxencoder$ is responsible for keeping track of the dialog context by
	encoding all features generated so far. 
The context vector at $t$ ($\mathbf{h}_t$) is updated using the historical 
    information from the previous step:
\begin{align}
	\boldv_{t - 1} & =  \squarex{\hspeaker_{t - 1}; 
								  \hgoal_{t - 1}; \hstate_{t - 1}; 
						    	  \hutt_{t - 1}} \nonumber \\
	\boldh_t & = \ctxencoder\roundx{\boldh_{t - 1}, 
	                                 \boldv_{t - 1}} \nonumber
\end{align}
where $\boldv_t$ is represents all features at the step $t$.

VHDA uses the context information to successively generate turn-level latent 
	variables using a series of generator networks:
\begin{align}
	\posteriormodel{\zspeaker_t}{\boldh_t, \zconv} & =
		\gaussian{\muspeaker_t}{\sigmaspeaker_t\identity} \nonumber \\
	\posteriormodel{\zgoal_t}{\boldh_t, \zconv, \zspeaker_t} & =
		\gaussian{\mugoal_t}{\sigmagoal_t\identity} \nonumber \\
	\posteriormodel{\zstate_t}{\boldh_t, \zconv, \zspeaker_t, \zgoal_t} & =
		\gaussian{\mustate_t}{\sigmastate_t\identity} \nonumber \\
	\posteriormodel{\zutt_t}{\boldh_t, \zconv, \zspeaker_t, 
	                         \zgoal_t, \zstate_t} & =
		\gaussian{\muutt_t}{\sigmautt_t\identity} \nonumber 
\end{align}
where all latents are assumed to be Gaussian.
In addition, we assume the standard Gaussian for the global latents: 
	$p\roundx{\zconv} = \gaussian{0}{\identity}$.
We implemented the Gaussian distribution encoders ($\mu$ and $\sigma$) 
    using fully-connected networks $f$.
We also apply $\softplus$ on the output of the networks to infer the 
	variance of the distributions.
Employing the reparameterization trick \cite{kingma2013auto} allows 
    standard backpropagation during training of our model.

\textbf{Approximate Posterior Networks}. 
We use a separate set of parameters $\phi$ and encoders to approximate
	the posterior distributions of latent variables from the evidence.
In particular, the model infers the global latents $\zconv$ using the 
    conversation encoder $\seqencoder^{(c)}$ solely from the 
    linguistic features:
$$
	\posteriorapp{\zconv}{\boldh^{(u)}_1, \ldots, \boldh^{(u)}_T} = 
		\gaussian{\mu^{(c)}}{\sigma^{(c)}\identity}.
$$
Similarly, the approximate posterior distributions of 
	all turn-level latent variables are estimated from the evidence in cascade, 
	while maintaining the global conditioning:
\begin{align}
	\posteriorapp{\zspeaker_t}{\boldh_t, \zconv, \boldh^{(r)}_t} & =
		\gaussian{\mu^{(r')}_t}{\sigma^{(r')}_t\identity} \nonumber \\
	\posteriorapp{\zgoal_t}{\boldh_t, \zconv,
	                        \boldz^{(r)}_t, \boldh^{(g)}_t} & =
		\gaussian{\mu^{(g')}_t}{\sigma^{(g')}_t\identity} \nonumber \\
	\posteriorapp{\zstate_t}{\boldh_t, \ldots,
	                         \boldz^{(g)}_t, \boldh^{(s)}_t} & =
		\gaussian{\mu^{(s')}_t}{\sigma^{(s')}_t\identity} \nonumber \\
	\posteriorapp{\zutt_t}{\boldh_t, \ldots, 
						   \boldz^{(s)}_t, \boldh^{(u)}_t} & =
		\gaussian{\mu^{(u')}_t}{\sigma^{(u')}_t\identity}, \nonumber
\end{align}
where all Gaussian parameters are estimated using fully-connected layers, 
	parameterized by $\phi$.

\textbf{Realization Networks}.
A series of generator networks successively decodes 
	dialog features from their respective latent spaces to realize
	the surface forms:
\begin{align}
	\posteriormodel{\boldr_t}{\boldh_t, \zconv, \zspeaker_t} & =
		\textstyle \decoder^{\roundx{r}}_\theta
					\roundx{\boldh_t, \zconv, \zspeaker_t} 
		\nonumber \\
	\posteriormodel{\boldg_t}{\boldh_t, \ldots, \zgoal_t} & =
	\textstyle \decoder^{\roundx{g}}_\theta
				\roundx{\boldh_t, \ldots, \zgoal_t} 
	\nonumber \\
	\posteriormodel{\bolds_t}{\boldh_t, \ldots, \zstate_t} & =
	\textstyle \decoder^{\roundx{s}}_\theta
				\roundx{\boldh_t, \ldots, \zstate_t} 
	\nonumber \\
	\posteriormodel{\boldu_t}{\boldh_t, 
								\ldots, \zutt_t} & =
		\textstyle \decoder^{\roundx{u}}_\theta
			\roundx{\boldh_t, \dots, \zutt_t}. \nonumber
\end{align}
The utterance decoder $\textstyle \decoder^{\roundx{u}}$ is implemented 
	using the LSTM cell.
To alleviate sparseness in goals and turn-level dialog acts, 
	we formulate the classification 
	problem as a set of binary classification problems \cite{mrkvsic2017neural}.
Specifically, given a candidate dialog act $a$, 
\begin{align}
	\posteriormodel{a\in\bolds_t}{\boldv_{<t}, \ldots} & =
		\sigma \roundx{\mathbf{o}^{(s)}_t
		               \cdot\textstyle\seqencoder^{(a)}\roundx{a}}
		\nonumber
\end{align}
where $\sigma$ is the sigmoid function and 
	$\mathbf{o}^{(s)}_t \in \mathbb{R}^{d^{(a)}}$ is the output of a 
	feedforward network parameterized by $\theta$ that predicts the dialog act 
	specification embeddings.
Goals are predicted analogously.

\subsection{Training Objective}

Given all the latent variables $\mathbf{z}$ in our model, we optimize
	the evidence lower-bound (ELBO) of the goal-oriented dialog samples 
	$\boldc$:
\begin{align}
	\mathcal{L}_{\text{VHDA}} = & \mathbb{E}_{q_\phi}
		[ \log p_\theta ( \boldc \mid \boldz)] \nonumber \\
		& - D_{\text{KL}} ( q_\phi ( \boldz \mid \boldc ) \| p ( \boldz )).
	\label{eq:vhda-elbo}
\end{align}
The reconstruction term of \autoref{eq:vhda-elbo} can be factorized into 
	posterior probabilities in the realization networks.
Similarly, the KL-divergence term can be factorized and reformulated in
	approximate posterior networks and conditional priors based on the
	graphical structure.

\subsection{Minimizing Inference Collapse}
\label{sec:vhda-vae}

Inference collapse is a relatively common phenomenon among autoregressive
	VAE structures ~\cite{zhao2017infovae}.
The hierarchical and recurrent nature of our model makes it especially 
	vulnerable.
The standard treatment for alleviating the inference collapse problem include
	(1) annealing the KL-divergence term weight during the initial training
	stage and (2) employing word dropouts on the decoder inputs
	 ~\cite{bowman2016generating}. 
For our model, we observe that the basic techniques are insufficient 
	(\autoref{tab:vhda-ablation}).
While more recent treatments exist ~\cite{kim2018semi, he2019lagging},
	they incur high computational costs that prohibit practical 
	deployment in our cases.
We introduce two simpler but effective methods to prevent encoder 
	degeneration.

\noindent\textbf{Mutual Information Maximization}. 
The KL-divergence term in the standard VAE ELBO can be 
	decomposed to reveal the mutual information term~\cite{hoffman2016elbo}:
\begin{align}
	\mathbb{E}_{p_d} [ D_{\text{KL}} &
		( q_\phi ( \boldz \mid \boldx ) \| p (\boldz)) ] = \nonumber \\
		& D_{\text{KL}} ( q_\phi(\boldz) \| p(\boldz) ) 
		+ I_{q_\phi} (\boldx; \boldz) \nonumber
\end{align} 
where $p_d$ is the empirical distribution of the data.
Re-weighting the decomposed terms for optimizing the VAE behaviors has been 
	explored previously 
	~\cite{chen2018isolating, zhao2017infovae, tolstikhin2018wasserstein}.
In this work, we propose simply canceling out the mutual information 
	term by performing mutual information estimation as a post-procedure.
Since the preservation of the conversation encoder $\seqencoder^{(c)}$ and
	global latents is vital for generation controlability, we 
	specifically maximize mutual information between the global latents and 
	the evidence:
\begin{align}
	\mathcal{L}_{\text{VHDA}} = & \mathbb{E}_{q_\phi}
		[ \log p_\theta ( \boldc \mid \boldz)] \label{eq:vhda-mim-elbo} \\
	& - D_{\text{KL}} ( q_\phi ( \boldz \mid \boldc ) \| p ( \boldz )) 
		+ I_{q_\phi} (\boldc; \zconv). \nonumber
\end{align}
In our work, the mutual information term is computed empirically using
	the Monte-Carlo estimator for each mini-batch.
The details are provided in the supplementary material.

\textbf{Hierarchically-scaled Dropout}.
Extending word dropouts and utterance dropouts \citet{park2018hierarchical}, 
	we apply dropouts discriminatively to all dialog features 
	(goals and dialog acts) according to the feature hierarchy level.
We hypothesize that employing dropouts could be detrimental to the 
	learning of lower-level latent variables, as information dropouts stack
	multiplicatively along the hierarchy.
However, it is also necessary in order to encourage meaningful encoding of
	latent variables.
Specifically, we propose a novel dropout scheme that scales exponentially along
	with the hierarchical depth, allowing higher-level information to
	flow towards lower levels easily.
For our implementation, we set the dropout ratio between two adjacent levels 
	to 1.5, resulting in the dropout probabilities of 
	[0.1, 0.15, 0.23, 0.34, 0.51] for speaker information to utterances.
We confirm our hypothesis in \autoref{sec:vhda-exp-gda}.

\section{Experiments}
\label{sec:vhda-experiments}

\subsection{Experimental Settings}

Following the protocol in ~\cite{yoo2019data},
	we generate three independent sets of synthetic dialog samples, and, for 
	each augmented dataset, we repeatedly train the same dialog state tracker
	three times with different seeds.
We compare the aggregated results from all nine trials with the baseline results.
Ultimately, we repeat this procedure for all combinations of state trackers and datasets. For non-augmented baselines, we repeat the experiments ten times.

\noindent\textbf{Implementation Details}. 
The hidden size of dialog vectors is 1000, and the hidden size of utterance, 
    dialog act specification, turn state, and turn goal representations is 500. 
The dimensionality for latent variables is between 100 and 200.
We use GloVe \cite{pennington2014glove} and character \cite{hashimoto2017joint} 
	embeddings as pre-trained word emebddings (400 dimensions total) 
	for word and dialog act tokens.
All models used Adam optimizer \cite{kingma2014adam} with the initial 
	learning rate of 1e-3, 
We annealed the KL-divergence weights over 250,000 training steps.
For data synthesis, we employ ancestral sampling to generate samples from the 
	empirical posterior distribution.
We fixed the ratio of synthetic to original data samples to 1.

\noindent\textbf{Datasets}. 
We conduct experiments on four state tracking corpora: 
	\textit{WoZ2.0} \cite{wen2017network},
	\textit{DSTC2} \cite{henderson2014second}, 
	\textit{MultiWoZ} \cite{budzianowski2018multiwoz},
	and \textit{DialEdit} \cite{manuvinakurike2018dialedit}.
These corpora cover a variety of domains
	(restaurant booking, hotel reservation, and image editing).
Note that, because the MultiWoZ dataset is a multi-domain corpus, we extract 
	single-domain dialog samples from the two most prominent domains
	(hotel and restaurant, denoted by \textit{MultiWoZ-H} and 
	\textit{MultiWoZ-R}, respectively).

\noindent\textbf{Dialog State Trackers}.
We use GLAD and GCE as the two competitive baselines for state tracking.
Besides, modifications are applied to these trackers 
	to stabilize the performance on random seeds (denoted as 
	GLAD\super{+} and GCE\super{+}).
Specifically, we enrich the word embeddings with subword information 
	~\cite{bojanowski2017enriching} and apply dropout on word embeddings 
	(dropout rate of 0.2).
Furthermore, we also conduct experiments on a simpler
	architecture that shares a similar structure with GCE but does not employ 
	self-attention for the sequence encoders (denoted as RNN).

\noindent\textbf{Evaluation Measures}.
\textit{Joint goal accuracy} (\textit{goal} for short) measures the ratio
	of the number of turns whose goals a tracker has correctly identified over the 
	total number of turns.
Similarly, \textit{request accuracy}, or \textit{request}, measures
	the turn-level accuracy of request-type dialog acts, while 
	\textit{inform accuracy} (\textit{inform}) measures the turn-level accuracy
	of inform-type dialog acts.
Turn-level goals accumulate from inform-type dialog acts starting from the 
	beginning of the dialog until respective dialog turns, and thus they can
	be inferred from historical inform-type dialog acts 
	(\autoref{tab:vhda-interpolation-2}).

\begin{table*}[t]
    \small
    \scshape
	\centering
	\begin{threeparttable}
		\begin{tabular}{llcccccccccc}
			\toprule
			\multirow{2}{*}{GDA} & 
			\multirow{2}{*}{Model} & 
			\multicolumn{2}{c}{WoZ2.0} & 
			\multicolumn{2}{c}{DSTC2} & 
			\multicolumn{2}{c}{MWoZ-R} & 
			\multicolumn{2}{c}{MWoZ-H} &
			\multicolumn{2}{c}{DialEdit} \\
			\cmidrule{3-12}
			& & Goal & Req & Goal & Req & Goal & Inf & 
				Goal & Inf & Goal & Req \\
			\midrule
			- & RNN &
				74.5 & 96.1 & 69.7 & 96.0 &
				43.7 & 69.4 & 25.7 & 55.6 &
				35.8 & 96.6 \\  
			VHDA & RNN &
				\textbf{78.7}\tnote{\ddag} & 
				\textbf{96.7}\tnote{\ddag} & 
				\textbf{74.2}\tnote{\dag} & 
				\textbf{97.0}\tnote{\ddag} &
				\textbf{49.6}\tnote{\dag} & 
				\textbf{73.4}\tnote{\dag} & 
				\textbf{31.0}\tnote{\dag} & 
				\textbf{59.7}\tnote{\dag} &
				\textbf{36.4}\tnote{\dag} & 
				\textbf{96.8} \\
			\midrule
			- & GLAD\textsuperscript{+} &
				87.8 & \textbf{96.8} & 
				74.5 & 96.4 &
				58.9 & 76.3 & 
				33.4 & 58.9 &
				35.9 & 96.7 \\  
			VHDA & GLAD\textsuperscript{+} &
				\textbf{88.4} & 
				96.6 & 
				\textbf{75.5}\tnote{\ddag} & 
				\textbf{96.8}\tnote{\dag} &
				\textbf{61.5}\tnote{\dag} & 
				\textbf{77.4} & 
				\textbf{37.8}\tnote{\ddag} & 
				\textbf{61.3}\tnote{\ddag} & 
				\textbf{37.1}\tnote{\dag} & 
				\textbf{96.8} \\
			\midrule
			- & GCE\textsuperscript{+} &
				88.7 & 97.0 & 74.8 & 96.3 &
				60.5 & 76.7 & 36.5 & 61.0 & 
				36.1 & 96.6 \\  
			VHDA & GCE\textsuperscript{+} &
				\textbf{89.3}\tnote{\ddag} & 
				\textbf{97.1} & 
				\textbf{76.0}\tnote{\ddag} & 
				\textbf{96.7}\tnote{\dag} &
				\textbf{63.3} & 
				\textbf{77.2} & 
				\textbf{38.3} & 
				\textbf{63.1}\tnote{\dag} &
				\textbf{37.6}\tnote{\dag} & 
				\textbf{96.8} \\
			\bottomrule
		\end{tabular}
		\begin{tablenotes}
			\item[\dag]{\scriptsize $p < 0.1$}
			\item[\ddag]{\scriptsize $p < 0.01$}
		\end{tablenotes}
	\end{threeparttable}
	\caption{
		Results of data augmentation using VHDA for dialog state tracking 
		on various datasets and state trackers. Note that we report inform 
		accuracies for MultiWoZ datasets instead, as request-type prediction is
		trivial for those.
	}
	\label{tab:vhda-gda}
\end{table*}

\subsection{Data Augmentation Results}
\label{sec:vhda-exp-gda}

\begin{table}
    \small
    \scshape
	\centering
	\begin{tabular}{llcccccccc}
		\toprule
		\multirow{2}{*}{Goal} & 
		\multirow{2}{*}{DST} & 
		\multicolumn{2}{c}{WoZ2.0} & 
		\multicolumn{2}{c}{DSTC2} \\
		\cmidrule{3-6}
		& & Goal & Req & Goal & Req \\
		\midrule
		w/o & RNN &
			77.8 & 96.4 & 71.2 & \textbf{97.2} \\  
		w/ & RNN &
			\textbf{78.7} & \textbf{96.7} & \textbf{74.2} & 97.0 \\
		\weakline
		w/o & GLAD\textsuperscript{+} &
			86.5 & \textbf{96.9} & 74.7 & \textbf{97.0} \\  
		w/ & GLAD\textsuperscript{+} &
			\textbf{88.4} & 96.6 & \textbf{75.5} & 96.8 \\
		\weakline
		w/o & GCE\textsuperscript{+} &
			86.4 & 96.3 & 75.5 & 96.7 \\  
		w/ & GCE\textsuperscript{+} &
			\textbf{89.3} & \textbf{97.1} & \textbf{76.0} & \textbf{96.7} \\
		\bottomrule
	\end{tabular}
	\caption{
		Comparison of data augmentation results between VHDA with and
		without explicit goal tracking.
	}
	\label{tab:vhda-gda-nogoal}
\end{table}

\noindent\textbf{Main Results}. 
We present the data augmentation results in \autoref{tab:vhda-gda}.
The results strongly suggest that generative data augmentation for dialog state 
	tracking is a viable strategy for improving existing DST models without 
	modifying them, as improvements were observed at statistically significant 
	levels regardless of the tracker and dataset.

The margin of improvements was more significant for less expressive state 
    trackers (RNN) than the more expressive ones (GLAD\super{+} and GCE\super{+}).
Even so, we observed varying degrees of improvements (zero to two percent in joint
    goal accuracy) even for the more expressive trackers, suggesting that GDA is 
    effective regardless of downstream model expressiveness.

We observe larger improvement margins for inform-type dialog acts 
    (or subsequently goals) from comparing performances between 
    the dialog act types.
This observation is because request-type dialog acts are generally 
    more dependent on the user utterance in the same turn rather than 
    requiring resolution of long-term dependencies, as illustrated 
    in the dialog sample (\autoref{tab:vhda-interpolation-2}).
The observation supports our hypothesis that more diverse synthetic 
    dialogs can benefit data augmentation by exploring unseen dialog dynamics.

Note that the goal tracking performances have relatively high variances due 
    to the accumulative effect of tracking dialogs. 
However, as an additional benefit of employing GDA, we observe that synthetic 
    dialogs help stabilize downstream tracking performances on DSTC2 and 
    MultiWoZ-R datasets.

\noindent\textbf{Effects of Joint Goal Tracking}.
Since user goals can be inferred from turn-level inform-type dialog acts,
	it may seem redundant to incorporate goal modeling into our model.
To verify its effectiveness, we train a variant of VHDA, 
	where the model does not explicitly track goals.
The results (\autoref{tab:vhda-gda-nogoal}) show that VDHA without
	explicit goal tracking suffers in joint goal accuracy but performs better
	in turn request accuracy for some instances.
We conjecture that explicit goal tracking helps the model reinforce long-term 
    dialog goals; however, the model does so in the minor expense of 
    short-term state tracking (as evident from lower state tracking accuracy).

\begin{table}
	\small
	\scshape
	\centering
	\begin{tabular}{llccc}
		\toprule
		\multirow{2}{*}{Drop.} & 
		\multirow{2}{*}{Obj.} &
		\multirow{2}{*}{$\zconv$-KL} &
		\multicolumn{2}{c}{WoZ2.0} \\
		\cmidrule{4-5}
		& & & Goal & Req \\
		\midrule
		0.00 & Std. & 5.63 & 84.1$\pm$0.9 & 95.9$\pm$0.6 \\
		0.00 & MIM  & 5.79 & 86.0$\pm$0.2 & 96.1$\pm$0.2 \\
		\weakline
		0.25 & Std. & 10.44 & 88.5$\pm$1.4 & 96.9$\pm$0.1 \\
		0.25 & MIM  & 11.31 & 88.9$\pm$0.4 & 97.0$\pm$0.2 \\
		\weakline
		0.50 & Std. & 14.68 & 88.6$\pm$1.0 & 96.9$\pm$0.2 \\
		0.50 & MIM  & 16.33 & 89.2$\pm$0.8 & 96.9$\pm$0.2 \\
		\weakline
		Hier. & Std. & 14.34 & 88.2$\pm$1.0 & 97.1$\pm$0.2 \\
		Hier. & MIM  & 16.27 & \textbf{89.3}$\pm$0.4 & \textbf{97.1}$\pm$0.2 \\
		\bottomrule
	\end{tabular}
	\caption{Ablation studies on the training techniques using 
             GCE\textsuperscript{+} as the tracker. The effect of different
             dropout schemes and training objectives is quantified. MIM 
             refers to mutual information maximization 
             (\autoref{sec:vhda-vae}).}
	\label{tab:vhda-ablation}
\end{table}

\noindent\textbf{Effects of Employing Training Techniques}.
To demonstrate the effectiveness of the two proposed training techniques,
	we compare (1) the data augmentation results and (2) the KL-divergence 
	between the posterior and prior of the dialog latents $\zconv$ 
	(\autoref{tab:vhda-ablation}).
The results support our hypothesis that the proposed 
	measures reduce the risk of inference collapse.
We also confirm that exponentially-scaled dropouts are more or comparably
	effective at preventing posterior collapse than uniform dropouts 
	while generating more coherent samples 
	(evident from higher data augmentation results).

\subsection{Language Evaluation}

\begin{table}
    \small
    \scshape
	\centering
	\begin{threeparttable}
		\begin{tabular}{lcccc}
			\toprule
			\multirow{2}{*}{Model} &
			\multicolumn{2}{c}{WoZ2.0} &
			\multicolumn{2}{c}{DSTC2} \\
			\cmidrule{2-5}
			& \scriptsize ROUGE & Ent & 
			\scriptsize ROUGE & Ent \\
			\midrule
			VHCR\super{\normalfont a} & 
				0.476 & 0.193 & 0.680 & 0.153 \\
			VHDA\super{\normalfont b} w/o goal & 
				0.473 & \textbf{0.195} & 0.743 & \textbf{0.162} \\
			VHDA\super{b} &
				\textbf{0.499} & 0.193 & 
				\textbf{0.781} & 0.154 \\
			\bottomrule
		\end{tabular}
		\begin{tablenotes}
			\item[\normalfont a]
				{\footnotesize\normalfont \cite{park2018hierarchical}}
			\item[\normalfont b]{\footnotesize\normalfont Ours}
		\end{tablenotes}
	\end{threeparttable}
	\caption{Results on language quality and diversity evaluation.}
    \label{tab:vhda-lang}
\end{table}

To understand the effect of joint learning of various dialog features on 
	language generation, we compare our model with a model that only learns
	linguistic features.
Following the evaluation protocol from prior work
	\cite{wen2017network, bak2019variational}, 
	we use ROUGE-L F1-score \cite{lin2004rouge} to
	evaluate the linguistic quality and utterance-level
	unigram cross-entropy \cite{serban2017hierarchical} (regarding the
	training corpus distribution) to evaluate diversity. 
\autoref{tab:vhda-lang} shows that our model generates better and more diverse
	utterances than the previous strong baseline on conversation 
	modeling.
These results supports the idea that joint learning of dialog annotations improves
	utterance generation, thereby increasing the chance of 
	generating novel samples that improve the downstream trackers.

\subsection{User Simulation Evaluation}

\begin{table}
    \small
    \scshape
	\centering
	\begin{threeparttable}
		\begin{tabular}{lcccc}
			\toprule
			\multirow{2}{*}{Model} &
			\multicolumn{2}{c}{WoZ2.0} &
			\multicolumn{2}{c}{DSTC2} \\
			\cmidrule{2-5}
			& Acc & Ent & Acc & Ent \\
			\midrule
			VHUS\super{\normalfont a} & 
				0.322 & 0.056 & 0.367 & 0.024 \\
			VHDA\super{\normalfont b} w/o GT & 
				0.408 & 0.079 & 0.460 & 0.034 \\
			VHDA\super{\normalfont b} &
				\textbf{0.460} & \textbf{0.080} & \textbf{0.554} & \textbf{0.043} \\
			\bottomrule
		\end{tabular}
		\begin{tablenotes}
			\item[\normalfont a]
				{\footnotesize\normalfont \cite{gur2018user}}
			\item[\normalfont b]{\footnotesize\normalfont Ours}
		\end{tablenotes}
	\end{threeparttable}
	\caption{Comparison of user simulation performances.}
    \label{tab:vhda-usersim}
\end{table}

Simulating human participants has become a crucial feature for training 
	dialog policy models using reinforcement learning and 
	automatic evaluation of dialog systems \cite{asri2016sequence}.
Although our model does not specialize in user simulation, our experiments show that
    the model outperforms the previous model (VHUS\footnote{The previous model employs 
	variational inference for contextualized sequence-to-sequence dialog act prediction.}) \cite{gur2018user} in terms
    of accuracy and creativeness (diversity).
We evaluate the user simulation quality using the prediction accuracy on the 
    test sets and the diversity using the entropy\footnote{The entropy is calculated with respect to the training set distribution} of predicted dialog act 
    specifications (\verb|act|-\verb|slot|-\verb|value| triples).
We present the results in \autoref{tab:vhda-usersim}.

\subsection{$\zconv$-interpolation}

\begin{table}
	\scriptsize
	\setlength\tabcolsep{1.5pt}
	\centering
	\begin{tabular}{
		p{0.02\linewidth}p{0.11\linewidth}
		>{\raggedright}p{0.27\linewidth}p{0.27\linewidth}p{0.24\linewidth}
	}
		\toprule
         & \textsc{Spkr.} & \textsc{Utterance} & 
           \textsc{Goal} & \textsc{Turn Act} \\
		\midrule
		1 & User & 
			i want to find a cheap restaurant in the north part of town . & 
			inform(area=north) \newline inform(price range=cheap) & 
			inform(area=north) \newline inform(price range=cheap) \\
		\weakline
		2 & Wizard & what food type are you looking for ? &  
			& request(slot=food)\\
		\weakline
		3 & User & any type of restaurant will be fine . & 
			inform(area=north) \newline inform(food=dontcare) \newline
			inform(price range=cheap) & 
			inform(food=dontcare) \\
		\weakline
		4 & Wizard & 
			the \verb|<place>| is a cheap indian restaurant in the north . 
			would you like more information ? & & \\
		\weakline
		5 & User & 
			what is the number ? & 
			inform(area=north) \newline inform(food=dontcare) \newline
			inform(price range=cheap) & request(slot=phone) \\
		\weakline
		6 & Wizard & 
			\verb|<place>| 's phone number is \verb|<number>| . 
			is there anything else i can help you with ? & & \\
		\weakline
		7 & User & no thank you . goodbye . & 
			inform(area=north) \newline inform(food=dontcare) \newline
			inform(price range=cheap) & \\
		\bottomrule
	\end{tabular}
	\caption{
		A sample generated from the midpoint between two latent variables
		in the $\zconv$ space encoded from two anchor data points.
	}
    \label{tab:vhda-interpolation-2}
\end{table}

We conduct $\zconv$-interpolation experiments to demonstrate that our model 
	can generalize the dataset space and learn to decode plausible 
	samples from unseen latent space.
The generated sample (\autoref{tab:vhda-interpolation-2}) 
	shows that our model can maintain coherence while generalizing
	key dialog features, such as the user goal and the dialog length.
As a specific example, given both anchors' user goals 
	(\verb|food=mediterranean| and \verb|food=indian|, respectively)
	\footnote{The supplementary material includes the full examples.}, 
	the generated midpoint between the two data points
	is a novel dialog with no specific food type
	(\verb|food=dontcare|).

\section{Conclusion}

We proposed a novel hierarchical and recurrent VAE-based architecture 
	to capture accurately the semantics of fully annotated goal-oriented 
	dialog corpora.
To reduce the risk of inference collapse while maximizing the generation 
	quality, we directly modified the training objective and devised a technique
	to scale dropouts along the hierarchy.
We showed that our proposed model VHDA was able 
	to achieve significant improvements for 
	various competitive dialog state trackers in diverse corpora through 
	extensive experiments.
With recent trends in goal-oriented dialog systems gravitating towards 
	end-to-end approaches \cite{lei2018sequicity},
	we wish to explore a self-supervised model, which discriminatively generates
	samples that directly benefit the downstream models for the target task.
We would also like to explore different implementations in 
	line with recent advances in dialog models, especially using large-scale
	pre-trained language models.

\section*{Acknowledgement}

We thank Hyunsoo Cho for his help with implementations 
    and Jihun Choi for the thoughtful feedback.
We also gratefully acknowledge support from Adobe Inc. in 
    the form of a generous gift to Seoul National University.

\bibliographystyle{acl_natbib}
\bibliography{anthology,emnlp2020}

\clearpage

\begin{appendices}


\section{Mutual Information Maximization for Mitigating Inference Collapse}

During the training of VAEs, inference collapse occurs when the model converges to
	a local optimum where the approximate posterior $q_\phi(\boldz \mid \boldx)$
	collapses to the prior $p (\boldz)$, indicating the vanishment of the
	encoder network due to the decoder's negligence of the encoder signals.
Quantifying, diagnosing, and devising a mitigation technique for
	the inference collapse phenomenon have been studied extensively 
	in the past \cite{chen2016variational, zhao2017infovae, cremer2018inference, 
					  razavi2019preventing, he2019lagging}.
However, current approaches for mitigating inference collapse are limited to
	significant modifications to the existing VAE framework 
	\cite{he2019lagging, kim2018semi} or limited to specific architectural
	designs \cite{razavi2019preventing}.
Current approaches do not work well on our model due to the complexity
	of our VAE structure.
Instead, we employ a relatively simple technique that directly modifies the 
	VAE objective.
By doing so, we mitigate any significant changes to the main VAE framework while 
	achieving satisfactory results on inference collapse mitigation.
Though not covered in this paper, our method has applications in other
	VAE structures.
In this appendix, we wish to delve more in-depth into the intuitions and detailed 
	implementation of our approach.

\textbf{Motivation}.
As first noted by \citet{hoffman2016elbo} (and subsequently
	utilized by \cite{zhao2017infovae, chen2018isolating}), the 
	KL-divergence term of the ELBO objective can be decomposed into
	two terms: (1) the KL-divergence between the aggregate posterior and 
	the prior and (2) the mutual information between the latent variables and 
	the data:
\begin{align}
	\mathbb{E}_{p_d} [ D_{\text{KL}} &
		( q_\phi ( \boldz \mid \boldx ) \| p (\boldz)) ] = \nonumber \\
	    & D_{\text{KL}} ( q_\phi(\boldz) \| p(\boldz) ) 
		+ I_{q_\phi} (\boldx; \boldz)
	\label{eq:kld-decompose}
\end{align}
where $p_d$ is the empirical distribution of data and the aggregate posterior 
	$q_\phi \roundx{\boldz}$ is obtained by marginalizing the approximate 
	posterior using the empirical distribution:
\begin{align}
	q_\phi \roundx{\boldz} = 
		\mathbb{E}_{\boldx \sim p_d} [ q_\phi \roundx{\boldz \mid \boldx} ].
\end{align}
Using the definition of inference collapse, we can deduce that the KL-divergence 
	term $D_{\text{KL}} ( q_\phi ( \boldz \mid \boldx ) \| p (\boldz))$ is zero
	during inference collapse. 
This fact implies that both decomposed terms in \autoref{eq:kld-decompose} must
	be zero since both terms are non-negative. 

\begin{figure*}
	\centering
	\includegraphics[width=0.7\textwidth]{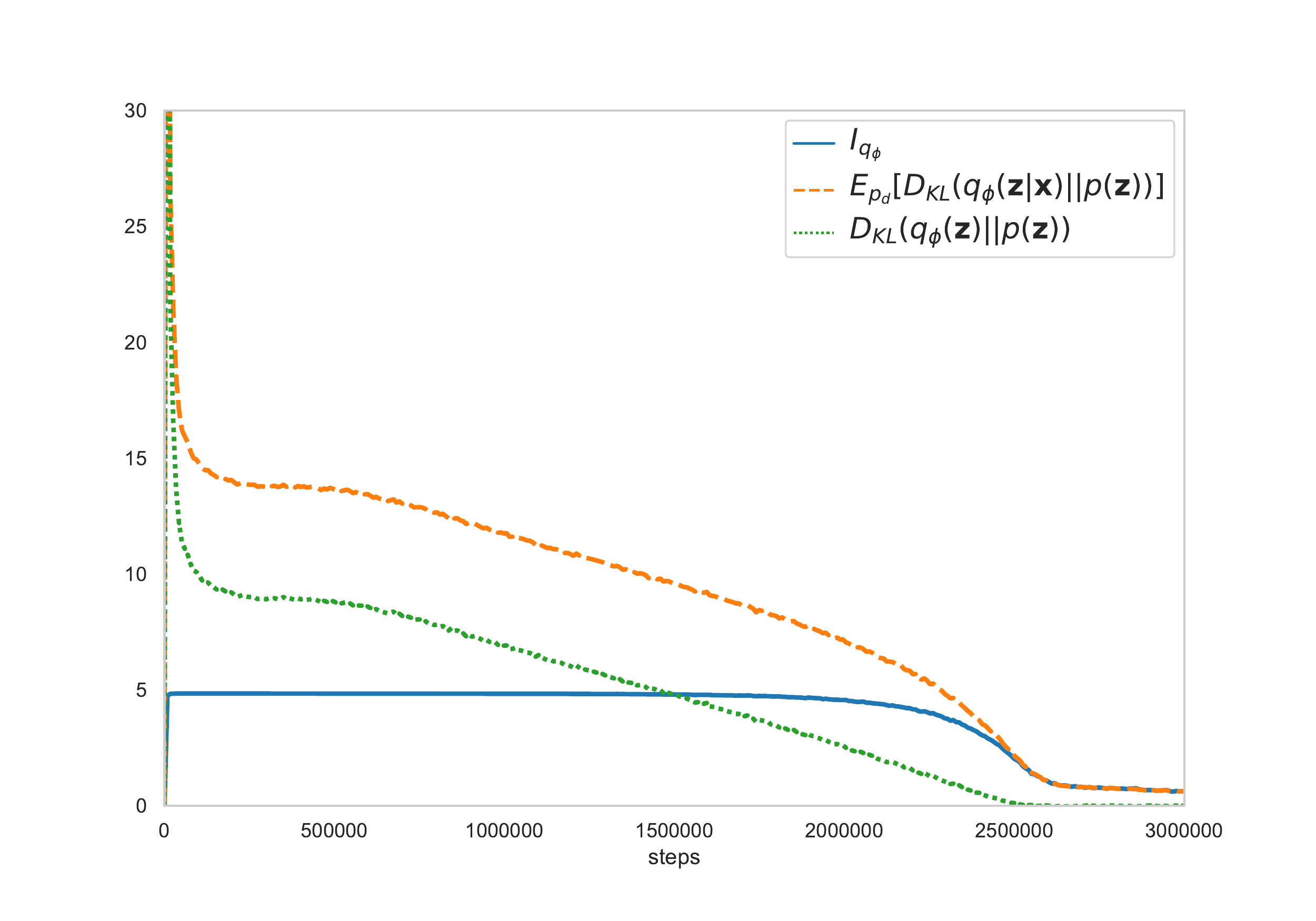}
	\caption{Failed training behavior.}
	\label{fig:vhda-fail}
\end{figure*}

Our preliminary studies show an interesting pattern in the KL-divergence term and its 
    decomposed terms during basic training (training without inference-collapse 
    treatments) (\autoref{fig:vhda-fail}).
We observe that the KL-divergence of the 
	aggregate posterior term vanishes earlier than the mutual information does.
We also observe that the mutual information term, which represents the encoder 
    effectiveness, vanishes eventually. 
This collapse happens after the KL-divergence cannot be minimized without 
    sacrificing the encoder's expressiveness.
Note that optimization of the ELBO objective minimizes the ELBO's KL-divergence 
	term and its underlying terms, one of which is directly related to the encoder
	health.
Although the reconstruction term in the ELBO encourages maximization of 
	the mutual information, the autoregressive property of the decoder and 
	the complexity of the reconstruction loss ``dilutes'' the goal of maximizing 
	mutual information.
Hence, to minimize inference collapse, we propose a modified VAE objective 
    that explicitly maximizes the mutual information between the latents and 
    the data by ``canceling" out the mutual information term in the KL-divergence\footnote{
		On a side note, we did not observe any ``lag'' in
		the inference network, as described by \citet{he2019lagging}.
		This observation is evident from the sustained mutual information level 
		throughout the training session (\autoref{fig:vhda-fail}).
		Hence we did not employ the recently proposed method.
	}:
\begin{align}
	\mathcal{L}_{\text{VHDA}} = 
		& \mathbb{E}_{p_d} [ \mathbb{E}_{q_\phi}
			[ \log p_\theta ( \boldc \mid \boldz) ] ]
		\nonumber \\ 
		& - \mathbb{E}_{p_d} [ 
			D_{\text{KL}} ( q_\phi ( \boldz \mid \boldc ) \| p ( \boldz )) ]
		\nonumber \\
		& + I_{q_\phi} (\boldc; \boldz).
	\label{eq:vhda-elbo}
\end{align}
Note that some notations (expectation over the empirical distribution) 
	have been omitted in the main paper for clarity.

\noindent\textbf{Relation to Prior Work}.
Our approach is related to previous work on manipulating the VAE objective
	for customizing the VAE behavior ~\cite{zhao2017infovae, chen2018isolating}.
It can also be thought of as a special case of Wasserstein Autoencoders 
	~\cite{tolstikhin2018wasserstein}
Although not all related works were original proposed to directly combat 
	inference collapse, our approach can be considered a special case of 
	InfoVAE ~\cite{zhao2017infovae} and $\beta$-TCVAE ~\cite{chen2018isolating}.
Specifically, ~\citet{zhao2017infovae} proposed a modified VAE 
	objective as follows:
\begin{align}
	\mathcal{L}_{\text{InfoVAE}} = 
		& \mathbb{E}_{p_d} [ \mathbb{E}_{q_\phi} [ 
			\log p_\theta (\boldx \mid \boldz) ]] \nonumber \\
		& - (1 - \alpha) \mathbb{E}_{p_d} [ D_{\text{KL}}
			( q_\phi ( \boldz \mid \boldx ) \| p (\boldz)) ] \nonumber \\
		& - (\alpha + \lambda  - 1) D_{\text{KL}} ( q_\phi(\boldz) \| p(\boldz) ).
	\label{eq:infovae}
\end{align}
Rearranging the equation, we can express the same objective related to the 
	mutual information:
\begin{align}
	\mathcal{L}_{\text{InfoVAE}} = 
		& \mathbb{E}_{p_d} [ \mathbb{E}_{q_\phi} [ 
			\log p_\theta (\boldx \mid \boldz) ]] \nonumber \\
		& - \lambda D_{\text{KL}} ( q_\phi(\boldz) \| p(\boldz) ) \nonumber \\
		& - (1 - \alpha) I_{q_\phi} (\boldx; \boldz).
	\label{eq:infovae}
\end{align}
Hence, our method is a special case of InfoVAE where $\alpha = 1$ and 
	$\lambda  =1$.
Meanwhile, \citet{chen2018isolating} proposed an extended 
	modification to $\beta$-VAE \cite{higgins2017beta} to further decompose
	the KL-divergence of the aggregate posterior in terms of latent correlation:
\begin{align}
	\mathcal{L}_{\text{InfoVAE}} = 
		& \mathbb{E}_{p_d} [ \mathbb{E}_{q_\phi} [ 
			\log p_\theta (\boldx \mid \boldz) ]] \nonumber \\
		& - \alpha I_{q_\phi} (\boldx; \boldz) \nonumber \\
		& - \beta D_{\text{KL}} ( q_\phi(\boldz) \| \textstyle\sum_i q_\phi(z_i) ) \nonumber \\
		& - \gamma \sum_i D_{\text{KL}} ( q_\phi(z_i) \| p(z_i) ).
	\label{eq:tcvae}
\end{align}
In the equation above, our approach corresponds the case where $\alpha = 0$ and $\beta = \gamma = 1$.

\textbf{Mutual Information Estimation}.
We can estimate the mutual information between the latents and the data under 
    the empirical distribution of $\boldx$ using Monte Carlo sampling.
However, this estimation method is known to be biased \cite{belghazi2018mine}.
Despite recent advances in MI estimation techniques, we
	find that our unparameterized method is sufficient for achieving inference
	collapse mitigation and probing.:
\begin{figure*}[b]
\begin{align}
	I_{q_\phi} \left( \boldx, \boldz \right) = & 
		\mathbb{E}_{p_d} \left[ 
			\kld{q_\phi \left( \boldz \mid \boldx \right)}
				{q_\phi \left( \boldz \right)} 
			\right] \nonumber \\
	\approx & \frac{1}{NM} \sum_i^N { 
		\sum_j^M { \left( 
			\log q_\phi( \boldz_j \mid \boldx_i )
			- \log \sum_k^L { q_\phi \left( \boldz_j \mid \boldx_k \right) } 
			+ \log L \right) 
		} 
	}
	\label{eq:mi-estimation}
\end{align}
\end{figure*}

\begin{figure*}
\begin{align}
	I_{q_\phi} \left( \boldx, \boldz \right) \approx & \frac{1}{N} 
		\sum_i^N { \squarex*{ \log { q_\phi \left( \boldz \mid \boldx_i \right) 
		- \log \sum_{j}^N { q_\phi \roundx*{\boldz \mid \boldx_j}} + 
						    \log N }}_{
								\boldz \sim q_\phi\roundx*{\boldz\mid\boldx_i}}}
	\label{eq:mi-estimation-2}
\end{align}
\end{figure*}

The equation for estimating the mutual information is shown in \autoref{eq:mi-estimation}.
where $\boldx$ is sampled from the empirical distribution of the dataset and
	$N$, $M$ and $L$ are hyperparameters.
In practice, the estimation is performed over the data samples in a mini-batch
	for computational efficiency.
Given a mini-batch of size $N$, we further approximate the estimation by 
	sampling the latent variables $\boldz$ once for each data point ($M = 1$) (\autoref{eq:mi-estimation-2}).

\begin{figure*}
	\centering
	\includegraphics[width=0.5\textwidth]{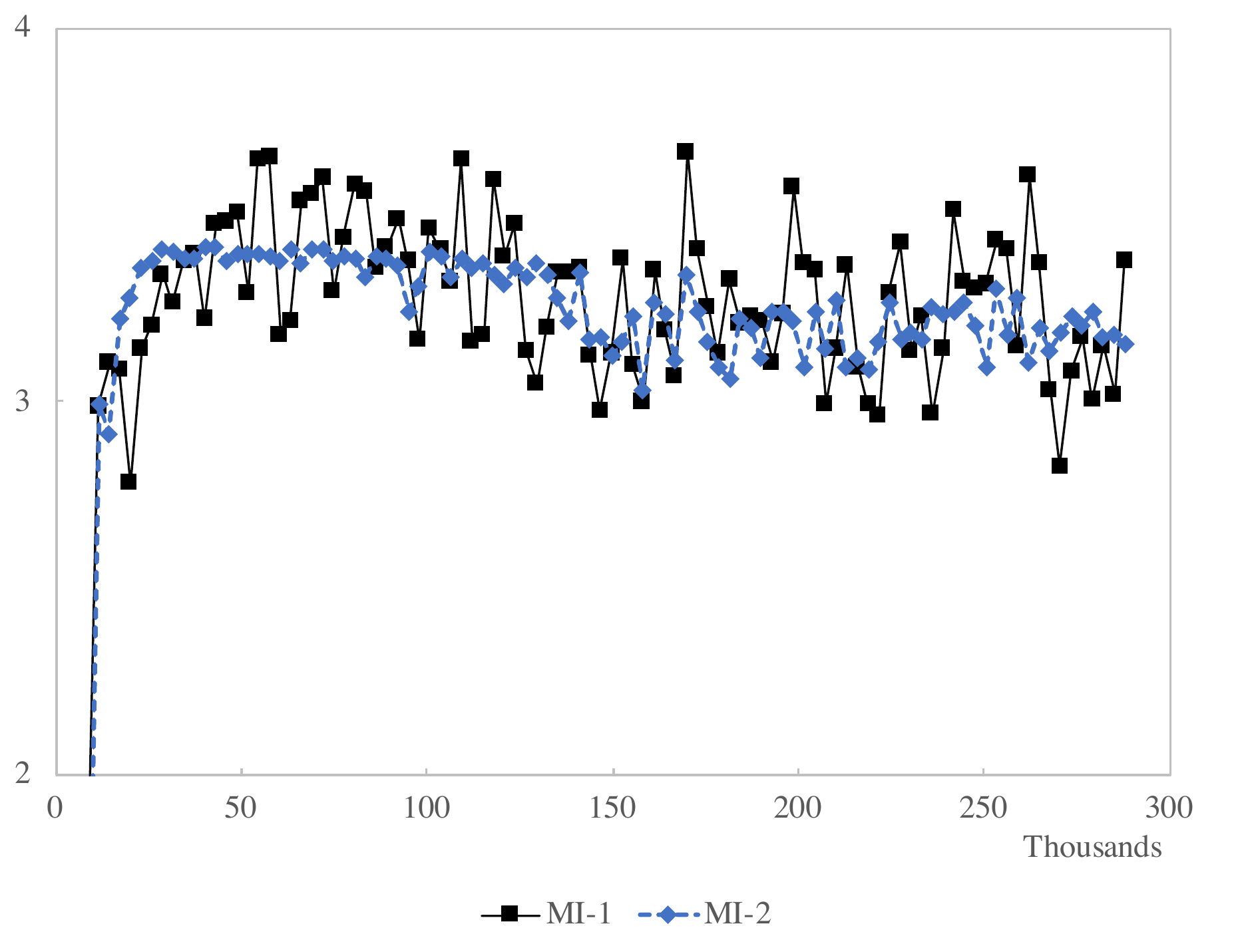}
	\caption{
		Estimation of the mutual information over the course of training.
		MI-2 corresponds to our approach. MI-1 is derived from the Monte
		Carlo estimation of $D_{\text{KL}} ( q_\phi(\boldz) \| p(\boldz) )$ 
		(not described). Our approach results in less variance in the 
		MI estimation.
	}
	\label{fig:mi-estimation}
\end{figure*}

We visualize the variance in our mutual information estimation method in 
	\autoref{fig:mi-estimation}.

\section{Architectural Diagram}

\begin{figure*}
	\centering
	\includegraphics[width=0.8\textwidth]{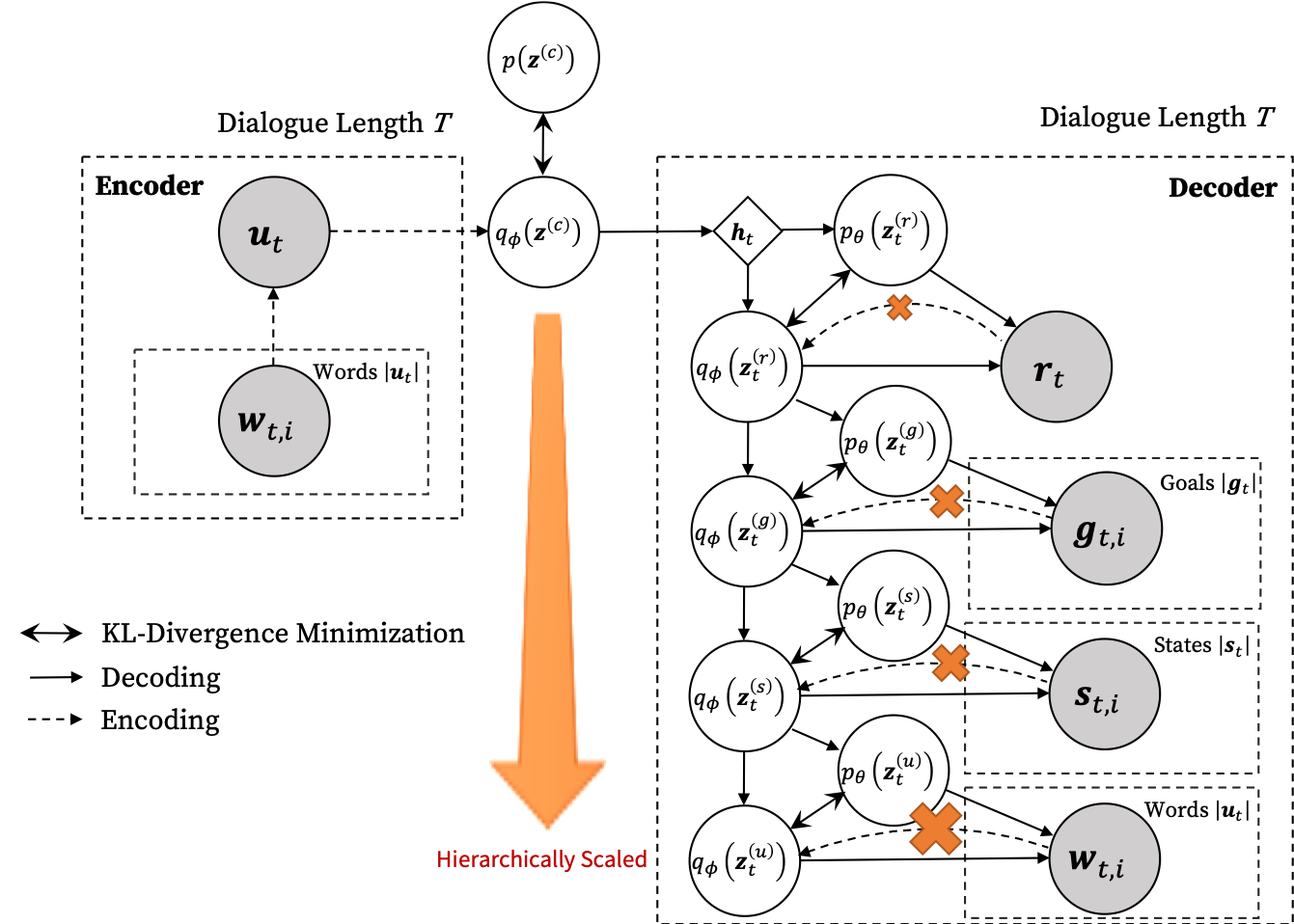}
	\caption{The architectural diagram.}
	\label{fig:app-architecture}
\end{figure*}

We include a more detailed architectural diagram (\autoref{fig:app-architecture})
    depicting the latent variables 
    and the model inference, which we could not illustrate in \autoref{fig:vhda}
    due to space constraints.
Note that the orange crosses denote decoder dropouts.
The figure also illustrates the hierarchically-scaled dropout scheme, 
	motivated by the need to minimize information loss while discouraging
	the decoders from relying on training signals, leading to exposure bias.

\begin{landscape}
\section{Full Results of Data Augmentation}

The full data augmentation results are shown below, including the statistics.
Note that generative data augmentation also has the effect of reducing the variance 
    of the downstream models.
\begin{table}[h]
	\vskip 1in
	\scshape
	\centering
	\begin{threeparttable}
		\begin{tabular}{llcccccccccc}
			\toprule
			\multirow{2}{*}{GDA} & 
			\multirow{2}{*}{Model} & 
			\multicolumn{2}{c}{WoZ2.0} & 
			\multicolumn{2}{c}{DSTC2} & 
			\multicolumn{2}{c}{MWoZ-R} & 
			\multicolumn{2}{c}{MWoZ-H} &
			\multicolumn{2}{c}{DialEdit} \\
			\cmidrule{3-12}
			& & Goal & Req & Goal & Req & Goal & Inf & 
				Goal & Inf & Goal & Req \\
			\midrule
			- & RNN &
				74.5$\pm$0.8 & 96.1$\pm$0.3 & 69.7$\pm$7.2 & 96.0$\pm$0.4 &
				43.7$\pm$8.7 & 69.4$\pm$5.7 & 25.7$\pm$4.1 & 55.6$\pm$2.3 &
				35.8$\pm$3.1 & 96.6$\pm$0.5 \\  
			VHDA & RNN &
				\textbf{78.7}$\pm$2.1\tnote{\ddag} & 
				\textbf{96.7}$\pm$0.1\tnote{\ddag} & 
				\textbf{74.2}$\pm$0.9\tnote{\dag} & 
				\textbf{97.0}$\pm$0.2\tnote{\ddag} &
				\textbf{49.6}$\pm$3.1\tnote{\dag} & 
				\textbf{73.4}$\pm$1.8\tnote{\dag} & 
				\textbf{31.0}$\pm$5.0\tnote{\dag} & 
				\textbf{59.7}$\pm$3.1\tnote{\dag} &
				\textbf{36.4}$\pm$1.4\tnote{\dag} & 
				\textbf{96.8}$\pm$0.1 \\
			\midrule
			- & GLAD\textsuperscript{+} &
				87.8$\pm$0.8 & \textbf{96.8}$\pm$0.3 & 
				74.5$\pm$0.5 & 96.4$\pm$0.2 &
				58.9$\pm$2.5 & 76.3$\pm$1.4 & 
				33.4$\pm$2.4 & 58.9$\pm$1.5 &
				35.9$\pm$1.0 & 96.7$\pm$0.3 \\  
			VHDA & GLAD\textsuperscript{+} &
				\textbf{88.4}$\pm$0.3 & 
				96.6$\pm$0.2 & 
				\textbf{75.5}$\pm$0.5\tnote{\ddag} & 
				\textbf{96.8}$\pm$0.5\tnote{\dag} &
				\textbf{61.5}$\pm$2.4\tnote{\dag} & 
				\textbf{77.4}$\pm$2.0 & 
				\textbf{37.8}$\pm$2.2\tnote{\ddag} & 
				\textbf{61.3}$\pm$1.0\tnote{\ddag} & 
				\textbf{37.1}$\pm$1.1\tnote{\dag} & 
				\textbf{96.8}$\pm$0.4 \\
			\midrule
			- & GCE\textsuperscript{+} &
				88.3$\pm$0.7 & 97.0$\pm$0.2 & 74.8$\pm$0.6 & 96.3$\pm$0.2 &
				60.5$\pm$3.4 & 76.7$\pm$1.2 & 36.5$\pm$2.4 & 61.0$\pm$1.2 & 
				36.1$\pm$1.3 & 96.6$\pm$0.4 \\  
			VHDA & GCE\textsuperscript{+} &
				\textbf{89.3}$\pm$0.4\tnote{\ddag} & 
				\textbf{97.1}$\pm$0.2 & 
				\textbf{76.0}$\pm$0.2\tnote{\ddag} & 
				\textbf{96.7}$\pm$0.4\tnote{\dag} &
				\textbf{63.3}$\pm$3.9 & 
				\textbf{77.2}$\pm$3.3 & 
				\textbf{38.3}$\pm$4.1 & 
				\textbf{63.1}$\pm$1.4\tnote{\dag} &
				\textbf{37.6}$\pm$2.1\tnote{\dag} & 
				\textbf{96.8}$\pm$0.4 \\
			\bottomrule
		\end{tabular}
		\begin{tablenotes}
			\item[\dag]{\scriptsize $p < 0.1$}
			\item[\ddag]{\scriptsize $p < 0.01$}
		\end{tablenotes}
	\end{threeparttable}
	\caption{The full results of data augmentation, including the standard 
				deviations of 9 repeated runs.}
	\label{tab:app-vhda-gda}
\end{table}
\end{landscape}

\pagebreak

\onecolumn

\section{Exhibits of Synthetic Samples}

This section describes the method we use to sample synthetic data 
	points from our model's posterior and presents some synthetic samples 
	generated from our model using the described technique.

We use \textit{ancestral sampling} \cite{he2019lagging}, or 
	the \textit{posterior sampling} technique \cite{yoo2019data},
	to sample data points from the empirical distribution of the latent space.
Specifically, we choose an anchor data point from the dialog dataset: 
	$\boldc \sim p_d\roundx{\boldc}$, where $p_d$ is the empirical 
	distribution of goal-oriented dialogs.
Then, we sample a set of latent variables $\zconv$ from the encoded distribution 
	of $\boldc$: $\zconv \sim q_\phi \roundx{\zconv\mid\boldc}$.
Next, we decode a sample $\boldc'$ that maximizes the log-likelihood 
	for each sampled conversational latents:
\begin{equation}
	\boldc' = \argmax_\boldc p_\theta \roundx{\boldc\mid\zconv}. \nonumber
\end{equation}
We use these samples to augment the original dataset.
Also, we fix the ratio of the synthetic dataset to the original dataset to 1.
In our experiments, we observe that all of the synthetic samples generated 
	via ancestral sampling are mostly coherent and, most importantly, novel, i.e.,
	each synthetic data point is somehow different from the original anchor 
	point (e.g., variations in utterances, dialog-level
	semantics, or sometimes annotation errors).
	
In the following tables, we showcase few dialog samples from our augmentation
	datasets.
The tables present the generated samples along with their reference dialog samples.

\vfill
{
	\footnotesize
	\setlength\tabcolsep{5pt}
	\centering
	\begin{tabular}{
		p{0.02\linewidth}p{0.07\linewidth}
		>{\raggedright}p{0.34\linewidth}p{0.24\linewidth}p{0.24\linewidth}
	}
		\toprule
			& \textsc{Speaker} & \textsc{Utterance} & 
			\textsc{Goal} & \textsc{Turn Act} \\
		\midrule
		\multicolumn{5}{c}{\scshape Anchor (Real)} \\
		\midrule    
		1 & User & i am looking for a panasian restaurant in the south side of town . if there are n't any maybe chinese . i need an address and price & inform(area=south) \newline inform(food=panasian) & inform(area=south) \newline inform(food=panasian) \newline request(slot=price range) \newline request(slot=address)\\
		\weakline
		2 & Wizard & there is an expensive and a cheap chinese restaurant in the south . which would you prefer ? &  & request(slot=price range)\\
		\weakline
		3 & User & let 's try cheap chinese restaurant . can i get an address ? & inform(area=south) \newline inform(food=chinese) \newline inform(price range=cheap) & inform(food=chinese) \newline inform(price range=cheap) \newline request(slot=address)\\
		\weakline
		4 & Wizard & of course it 's \verb|<location>| &  & \\
		\weakline
		5 & User & thank you goodbye . & inform(area=south) \newline inform(food=chinese) \newline inform(price range=cheap) & \\
		\midrule
		\multicolumn{5}{c}{\scshape Pointwise Posterior Sample (Generated)} \\
		\midrule
		1 & User & i 'm looking for a panasian restaurant in the south side of town . if there are n't any maybe chinese . i need an address and price & inform(food=panasian) \newline inform(area=south) & inform(food=panasian) \newline inform(area=south) \newline request(slot=price range)\\
		\weakline
		2 & Wizard & there are no cheap restaurants serving restaurants i have a seafood the the number is some other available &  & \\
		\weakline
		3 & User & how about thai & inform(food=thai) \newline inform(price range=cheap) & request(slot=address)\\
		\weakline
		4 & Wizard & we 's \verb|<place>| on \verb|<location>| &  & \\
		\weakline
		5 & User & thank you very much . & inform(food=thai) \newline inform(price range=cheap) & \\
		\bottomrule
	\end{tabular}
}

{
	\footnotesize
	\setlength\tabcolsep{4pt}
	\centering
	\begin{tabular}{
		p{0.02\linewidth}p{0.07\linewidth}
		>{\raggedright}p{0.34\linewidth}p{0.24\linewidth}p{0.24\linewidth}
	}
		\toprule
         & \textsc{Speaker} & \textsc{Utterance} & 
           \textsc{Goal} & \textsc{Turn Act} \\
		\midrule
		\multicolumn{5}{c}{\scshape Anchor (Real)} \\
		\midrule    
        1 & User & i need the address of a gastropub in town . & inform(food=gastropub) & inform(food=gastropub) \newline request(slot=address)\\
        \weakline
        2 & Wizard & which part of town ? &  & request(slot=area)\\
        \weakline
        3 & User & does n't matter . & inform(food=gastropub) \newline inform(area=dont care) & inform(area=dont care)\\
        \weakline
        4 & Wizard & would you prefer moderate or expensive pricing ? &  & request(slot=price range)\\
        \weakline
        5 & User & moderate please . & inform(food=gastropub) \newline inform(area=dont care) \newline inform(price range=moderate) & inform(price range=moderate)\\
        \weakline
        6 & Wizard & i have found one results that matches your criteria the restaurant the \verb|<place>| is a gastropub located at \verb|<location>| some \verb|code| as the price range is moderate &  & \\
        \weakline
        7 & User & are there any others in that price range ? & inform(food=gastropub) \newline inform(area=dont care) \newline inform(price range=moderate) & \\
        \weakline
        8 & Wizard & unfortunately there are not sorry &  & \\
        \weakline
        9 & User & hello i am looking for a restaurant that serves gastropub food in any area can you help me ? & inform(food=gastropub) \newline inform(area=dont care) \newline inform(price range=moderate) & \\
        \weakline
        10 & Wizard & sure would you prefer expensive or moderately priced ? &  & request(slot=price range)\\
        \weakline
		11 & User & thank you goodbye & inform(food=gastropub) \newline inform(area=dont care) \newline inform(price range=moderate) & \\
		\midrule
		\multicolumn{5}{c}{\scshape Pointwise Posterior Sample (Generated)} \\
		\midrule
		1 & User & i need the address of a gastropub in town . &  inform(food=gastropub) & inform(food=gastropub)\\
        \weakline
        2 & Wizard & i have many options . would you prefer centre or east ? &  & request(slot=area)\\
        \weakline
        3 & User & does n't matter . & inform(food=gastropub) \newline inform(area=dont care) \newline inform(area=center) & inform(area=dont care) \newline inform(area=center)\\
        \weakline
        4 & Wizard & there are three gastropub restaurants listed . one is in the east part of town and the rest are in the centre . &  & request(slot=price range)\\
        \weakline
        5 & User & i do n't care & inform(food=gastropub) \newline inform(price range=moderate) \newline inform(area=dont care) & inform(price range=moderate)\\
        \weakline
        6 & Wizard & i found \verb|<place>| . results that matches your criteria the restaurant the \verb|<place>| is a gastropub located at \verb|<location>| some \verb|<code>| as the price range is moderate &  & \\
        \weakline
        7 & User & are there any others in that price range ? & inform(food=gastropub) \newline inform(price range=moderate) \newline inform(area=dont care) & \\
        \weakline
        8 & Wizard & in that actually not sorry &  & \\
        \weakline
        9 & User & hello i am looking for a restaurant that serves gastropub food in any area can you help me ? & inform(food=gastropub) \newline inform(price range=moderate) \newline inform(area=dont care) & \\
		\bottomrule
	\end{tabular}
}

{
	\footnotesize
	\setlength\tabcolsep{4pt}
	\centering
	\begin{tabular}{
		p{0.02\linewidth}p{0.07\linewidth}
		>{\raggedright}p{0.34\linewidth}p{0.24\linewidth}p{0.24\linewidth}
	}
		\toprule
         & \textsc{Speaker} & \textsc{Utterance} & 
           \textsc{Goal} & \textsc{Turn Act} \\
		\midrule
		\multicolumn{5}{c}{\scshape Anchor (Real)} \\
		\midrule    
		1 & User & i 'm looking for a cheap restaurant in the west part of town . & inform(area=west) \newline inform(price range=cheap) & inform(area=west) \newline inform(price range=cheap)\\
        \weakline
        2 & Wizard & i found a vietnamese and italian cheap restaurant in the west side of town . would you like the phone number or address of either ? &  & request(slot=phone) \newline request(slot=address)\\
        \weakline
        3 & User & yes please . & inform(area=west) \newline inform(price range=cheap) & request(slot=phone) \newline request(slot=address)\\
        \weakline
        4 & Wizard & \verb|<place>| is the italian restaurant located at \verb|<location>| . its phone number is \verb|<numeric>| . \verb|<place>| is the vietnamese restaurant located at \verb|<location>| . its phone number is &  & \\
        \weakline
        5 & User & thank you . & inform(area=west) \newline inform(price range=cheap) & \\
        \weakline
        6 & Wizard & you 're welcome &  & \\
        \weakline
        7 & User & goodbye . & inform(area=west) \newline inform(price range=cheap) & \\
		\midrule
		\multicolumn{5}{c}{\scshape Pointwise Posterior Sample 1 (Generated)} \\
		\midrule
		1 & User & i 'm looking for a cheap restaurant in the west part of town . & inform(price range=cheap) \newline inform(area=west) & inform(price range=cheap)\\
        \weakline
        2 & Wizard & there is a cheap restaurant in the west part of town . would you like their address and location ? &  & request(slot=phone) \newline request(slot=address)\\
        \weakline
        3 & User & yes please . & inform(area=west) \newline inform(area=north) \newline inform(price range=cheap) & request(slot=phone) \newline request(slot=address)\\
        \weakline
        4 & Wizard & \verb|<place>| is the italian restaurant . &  & \\
		\weakline
		5 & User & thank you very much goodbye . & inform(area=north) \newline inform(price range=cheap) & \\
		\midrule
		\multicolumn{5}{c}{\scshape Pointwise Posterior Sample 2 (Generated)} \\
		\midrule
		1 & User & i want a cheap restaurant on the west side . & inform(price range=cheap) \newline inform(area=west) & inform(price range=cheap) \newline inform(area=west)\\
        \weakline
        2 & Wizard & \verb|<place>| is a restaurant that matches your choice in the west . &  & \\
        \weakline
        3 & User & \verb|<place>| the phone and the address ? & inform(food=vietnamese) \newline inform(price range=cheap) \newline inform(area=west) & request(slot=phone) \newline request(slot=address)\\
        \weakline
        4 & Wizard & \verb|<place>| 's phone number is \verb|<numeric>| &  & \\
        \weakline
        5 & User & thank you that will do . & inform(food=vietnamese) \newline inform(price range=cheap) \newline inform(area=west) & \\
		\bottomrule
	\end{tabular}
}

\pagebreak
\section{$\zconv$ Interpolation Results (Including Both Anchors)}

Visualizing samples from a linear interpolation of two points in the 
	latent space \cite{bowman2016generating} is a popular way to 
	showcase the generative capability of VAEs.
Given two dialog samples $\boldc_1$ and $\boldc_2$, we map the data points onto 
	the conversational latent space to obtain $\zconv_1$ and $\zconv_2$.
Multiple equidistant samples $\boldz'_1, ..., \boldz'_N$ are selected from the 
	linear interpolation between the two points: 
	$\boldz'_n = \zconv_1 + n(\zconv_2 - \zconv_1) / N$.
Likelihood-maximizing samples $\boldx'_1, \ldots, \boldx'_N$ are chosen from the 
	model posteriors given the intermediate latent samples.

{
	\renewcommand{\weakline}{
		\noalign{\global\arrayrulewidth=0.25\arrayrulewidth}
		\arrayrulecolor{lightgray}
		\\ [-2.4ex] \hline \\ [-2.4ex]
		\noalign{\global\arrayrulewidth=4.0\arrayrulewidth}
		\arrayrulecolor{black}
	}
	\footnotesize
	\setlength\tabcolsep{5pt}
	\centering
	\begin{tabular}{
		p{0.02\linewidth}p{0.07\linewidth}
		>{\raggedright}p{0.34\linewidth}p{0.24\linewidth}p{0.24\linewidth}
	}
		\toprule
         & \textsc{Speaker} & \textsc{Utterance} & 
           \textsc{Goal} & \textsc{Turn Act} \\
		\midrule
		\multicolumn{5}{c}{\scshape Anchor 1 (Real)} \\
		\midrule    
		1 & User & 
			i 'm looking for a mediterranean place for any price . 
			what is the phone and postcode ? & 
			inform(food=mediterranean) \newline inform(price=dont care) & 
			inform(food=mediterranean) \newline inform(price=dont care) 
			\newline request(slot=phone) \newline request(slot=postcode) \\
		\weakline
		2 & Wizard & 
			i found a few places . 
			the first is \verb|<place>| with a phone number of 
			\verb|<number>| and postcode of \verb|<postcode>| & & \\
		\weakline
		3 & User & 
			That will be fine . thank you . & 
			inform(food=mediterranean) \newline inform(price=dont care) & \\
		\midrule
		\multicolumn{5}{c}{\scshape Midpoint 50\% (Generated)} \\
        \midrule    
		1 & User & 
			i want to find a cheap restaurant in the north part of town . & 
			inform(area=north) \newline inform(price range=cheap) & 
			inform(area=north) \newline inform(price range=cheap) \\
		\weakline
		2 & Wizard & what food type are you looking for ? &  
			& request(slot=food)\\
		\weakline
		3 & User & any type of restaurant will be fine . & 
			inform(area=north) \newline inform(food=dontcare) \newline
			inform(price range=cheap) & 
			inform(food=dontcare) \\
		\weakline
		4 & Wizard & 
			the \verb|<place>| is a cheap indian restaurant in the north . 
			would you like more information ? & & \\
		\weakline
		5 & User & 
			what is the number ? & 
			inform(area=north) \newline inform(food=dontcare) \newline
			inform(price range=cheap) & request(slot=phone) \\
		\weakline
		6 & Wizard & 
			\verb|<place>| 's phone number is \verb|<number>| . 
			is there anything else i can help you with ? & & \\
		\weakline
		7 & User & no thank you . goodbye . & 
			inform(area=north) \newline inform(food=dontcare) \newline
			inform(price range=cheap) & \\
		\midrule
		\multicolumn{5}{c}{\scshape Anchor 2 (Real)} \\
		\midrule    
		1 & User & 
			i am looking for a cheap restaurant in the 
			north part of town . & 
			inform(area=north) \newline inform(price range=cheap) & 
			inform(area=north) \newline inform(price range=cheap) \\
		\weakline
		2 & Wizard & 
			there are two restaurants that fit your criteria 
			would you prefer italian or indian food ? &  & 
			request(slot=food) \\
		\weakline
		3 & User & 
			let s try indian please & 
			inform(area=north) \newline inform(price range=cheap) \newline 
			inform(food=indian) & 
			inform(food=indian) \\
		\weakline
		4 & Wizard & 
			\verb|<name>| serves indian food in the cheap price range 
			and in the north part of town . is there anything else 
			i can help you with ? & & \\
		\weakline
		5 & User & 
			what is the name of the italian restaurant ? & 
			inform(area=north) \newline inform(price range=cheap) \newline 
			inform(food=indian) & 
			inform(food=italian) \newline request(slot=name) \\
		\weakline
		6 & Wizard & \verb|<name>| & & \\
		\weakline
		7 & User & 
			what is the address and phone number ? & 
			inform(area=north) \newline inform(price range=cheap) \newline 
			inform(food=indian) & 
			request(slot=address) \newline request(slot=phone) \\
		\weakline
		8 & Wizard & 
			the address for \verb|<name>| is \verb|<address>| and 
			the phone number is \verb|<phone>| . & & \\
		\weakline
		9 & User & 
			thanks so much . & 
			inform(area=north) \newline inform(price range=cheap) \newline 
			inform(food=indian) & \\
		\bottomrule
	\end{tabular}
}

{
	\footnotesize
	\setlength\tabcolsep{5pt}
	\centering
	\begin{tabular}{
		p{0.02\linewidth}p{0.07\linewidth}
		>{\raggedright}p{0.34\linewidth}p{0.24\linewidth}p{0.24\linewidth}
	}
		\toprule
         & \textsc{Speaker} & \textsc{Utterance} & 
           \textsc{Goal} & \textsc{Turn Act} \\
		\midrule
		\multicolumn{5}{c}{\scshape Anchor 1 (Real)} \\
		\midrule    
        1 & User & hi i 'm looking for a moderately priced restaurant in the south part of town . & inform(area=south) \newline inform(price range=moderate) & inform(area=south) \newline inform(price range=moderate)\\
        \weakline
        2 & Wizard & the \verb|<place>| \verb|<location>| is moderately priced and in the south part of town . would you like their location ? &  & request(slot=address)\\
        \weakline
        3 & User & yes . i would like the location and the phone number please . & inform(area=south) \newline inform(price range=moderate) & request(slot=phone) \newline request(slot=address)\\
        \weakline
        4 & Wizard & the address of \verb|<place>| \verb|<location>| is \verb|<location>| and the phone number is \verb|<numeric>| . &  & \\
        \weakline
		5 & User & thank you goodbye . & inform(area=south) \newline inform(price range=moderate) & \\
		\midrule
		\multicolumn{5}{c}{\scshape 30\% (Generated)} \\
        \midrule
        1 & User & i am looking for some seafood what can you tell me ? & inform(area=dont care) & inform(food=seafood) \newline inform(area=dont care)\\
        \weakline
        2 & Wizard & \verb|<place>| restaurant bar serves mexican food in the south part of town . would you like their location ? &  & request(slot=address)\\
        \weakline
        3 & User & yes i 'd like the address phone number and postcode please . & inform(food=lebanese) \newline inform(food=seafood) & request(slot=address) \newline request(slot=phone)\\
        \weakline
        4 & Wizard & \verb|<place>| is located at \verb|<location>| cost the phone number is \verb|<numeric>| . &  & \\
        \weakline
        5 & User & thank you goodbye . & inform(food=seafood) \newline inform(area=dont care) & \\
		\midrule
		\multicolumn{5}{c}{\scshape 70\% (Generated)} \\
        \midrule
        1 & User & i would like to find a restaurant in the east part of town that serves gastropub food . & inform(food=mexican) & inform(food=mexican)\\
        \weakline
        2 & Wizard & \verb|<place>| restaurant bar serves mexican food in the south part of town . would you like their location ? &  & request(slot=address)\\
        \weakline
        3 & User & yes i 'd like the address phone number and postcode please . & inform(food=mexican) & request(slot=address) \newline request(slot=postcode) \newline request(slot=phone)\\
        \weakline
        4 & Wizard & \verb|<place>| restaurant bar is located at \verb|<location>| . the postal code is some code and the phone number is \verb|<numeric>| . &  & \\
        \weakline
        5 & User & thank you goodbye . & inform(food=mexican) & \\
		\midrule
		\multicolumn{5}{c}{\scshape Anchor 2 (Real)} \\
		\midrule    
        1 & User & i want to find a restaurant in any part of town and serves malaysian food . & inform(area=dont care) \newline inform(food=malaysian) & inform(area=dont care) \newline inform(food=malaysian)\\
        \weakline
        2 & Wizard & there are no malaysian restaurants . would you like something different ? &  & \\
        \weakline
        3 & User & north american please . give me their price range and their address and phone number please . & inform(area=dont care) \newline inform(food=north american) & inform(food=north american) \newline request(slot=phone) \newline request(slot=price range) \newline request(slot=address)\\
        \weakline
        4 & Wizard & \verb|<place>| is in the expensive price range their phone number is \verb|<numeric>| and their address is \verb|<location>| &  & \\
        \weakline
        5 & User & thank you goodbye & inform(area=dont care) \newline inform(food=north american) & \\
		\bottomrule
	\end{tabular}
}
\end{appendices}

\end{document}